\documentclass[11pt]{article}
\usepackage{geometry}
\usepackage{coling2020}
\usepackage{times}
\usepackage{url}
\usepackage{latexsym}
\usepackage{microtype}
\usepackage{graphicx}
\usepackage{xspace}
\usepackage{fontawesome}
\usepackage{float}

\usepackage{xcolor, colortbl}
\definecolor{Gray}{gray}{0.9}

\usepackage{tikz}
\usetikzlibrary{decorations.pathreplacing}

\DeclareGraphicsExtensions{.pdf,.ai,.jpg,.png}
\usepackage{lscape}	
\usepackage{booktabs}
\usepackage{todonotes}
\newcommand{\Ni}{({\em i})~}
\newcommand{\Nii}{({\em ii})~}

\newcommand{\corpus}{PTC-SemEval20 }
\newcommand{\sq}{\faCheckSquare}
\newcommand{\cq}{\faCheck}

\newcommand{\ranksi}[1]{{\scriptsize \texttt{(SI:#1)}}}
\newcommand{\ranktc}[1]{{\scriptsize \texttt{(TC:#1)}}}
\newcommand{\rankboth}[2]{{\scriptsize \texttt{(SI: #1, TC: #2)}}}

\newcommand{\pn}[1]{{ #1}}

\colingfinalcopy 

\title{SemEval-2020 Task 11:\\ Detection of Propaganda Techniques in News Articles}

\author{%
Giovanni Da San Martino$^1$, 
Alberto Barr\'{o}n-Cede\~no$^2$,\\
\textbf{Henning  Wachsmuth}$^3$,    
\textbf{Rostislav Petrov}$^4$ \and
\textbf{Preslav Nakov}$^1$   \\
  $^1$Qatar Computing Research Institute, HBKU, Qatar \hspace{5mm}
  $^2$Universit\`{a} di Bologna, Forl\`{i}, Italy \\
  $^3$Paderborn University, Paderborn, Germany \hspace{5mm}
  $^4$A Data Pro, Sofia, Bulgaria \\
  {\tt \{gmartino, pnakov\}@hbku.edu.qa}\hspace{5mm}
{\tt a.barron@unibo.it} \\
{\tt henningw@upb.de}\hspace{5mm}
{\tt rostislav.petrov@adata.pro}\\
}

\date{}

\begin{document}
\maketitle
\begin{abstract}
    We present the results and the main findings of SemEval-2020 Task 11 on Detection of Propaganda Techniques in News Articles. The task featured two subtasks. Subtask SI is about \emph{Span Identification}: given a plain-text document, spot the specific text fragments containing propaganda. Subtask TC is about \emph{Technique Classification}: given a specific text fragment, in the context of a full document, determine the propaganda technique it uses, choosing from an inventory of 14 possible propaganda techniques. The task attracted a large number of participants: \textbf{250} teams signed up to participate and \textbf{44} made a submission on the test set. In this paper, we present the task, analyze the results, and discuss the system submissions and the methods they used. For both subtasks, the best systems used pre-trained Transformers and ensembles.
\end{abstract}

\section{Introduction}
\label{intro}

%
%
\blfootnote{
    %
    %
    %
    %
 
    \hspace{-0.65cm}  
    This work is licensed under a Creative Commons 
    Attribution 4.0 International License.\\
    License details:
    \url{http://creativecommons.org/licenses/by/4.0/}.
}

Propaganda aims at influencing people's mindset with the purpose of advancing a specific agenda. 
It can hide in news published by both established and non-established outlets, and, in the Internet era, it has the potential of reaching very large audiences~\cite{Muller2018,brazil,Glowacki:18}. Propaganda is most successful when it goes unnoticed by the reader, and it often takes some training for people to be able to spot it.
The task is way more difficult for inexperienced users, and the volume of text produced on a daily basis makes it difficult for experts to cope with it manually. 
With the recent interest in ``fake news'', the detection of propaganda or highly biased texts has emerged as an active research area. However, most previous work has performed analysis at the document level only~\cite{Rashkin,AAAI2019:proppy} or has analyzed the general patterns of online propaganda~\cite{Garimella2015,Reddick2015}.

SemEval-2020 Task 11 offers a different perspective: a fine-grained analysis of the text that complements existing approaches and can, in principle, be combined with them. 
Propaganda in text (and in other channels) is conveyed through the use of diverse propaganda techniques~\cite{Miller}, 
which range from leveraging on the emotions of the audience ---such as using \textit{loaded language} or \textit{appeals to fear}--- to using logical fallacies ---such as \textit{straw men} (misrepresenting someone's opinion), hidden \textit{ad-hominem fallacies}, and \textit{red herring} (presenting irrelevant data). 
Some of these techniques have been studied in tasks such as hate speech detection \cite{I17-1078} and computational argumentation \cite{habernal:2018a}.

\begin{figure}[tbh]
    \centering
    \includegraphics[scale=1.0]{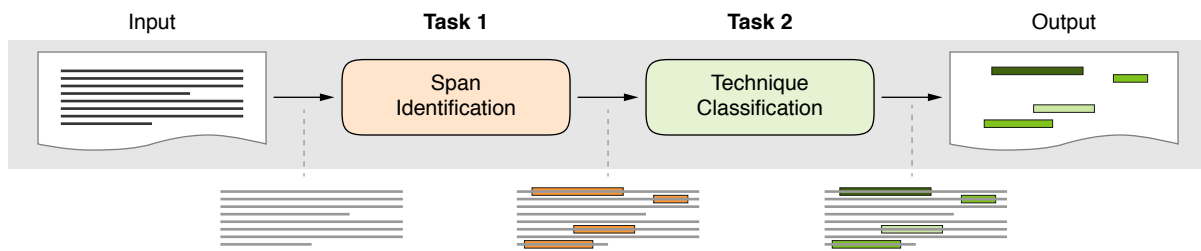}
    \caption{The full propaganda identification pipeline, including the two subtasks: Span Identification and Technique Classification. }
    \label{fig:pipeline}
\end{figure}

Figure~\ref{fig:pipeline} shows the fine-grained propaganda identification pipeline, including the two targeted subtasks. Our goal is to facilitate the development of models capable of spotting text fragments where propaganda techniques are used. The task featured the following subtasks:

\begin{description}
\item \textbf{Subtask SI} \textit{(Span Identification)}: Given a plain-text document, identify those specific fragments that contain at least one propaganda technique. This is a binary sequence tagging task. 
\item \textbf{Subtask TC} \textit{(Technique Classification)}: Given a text snippet identified as propaganda and its document context, identify the applied propaganda technique in the snippet. This is a multi-class classification problem. 
\end{description}

A total of $250$ teams registered for the task, $44$ of them made an official submission on the test set ($66$ submissions for both subtasks), and $32$ of the participating teams submitted a system description paper. 

The rest of the paper is organized as follows. 
Section~\ref{sec:propagandatechniques} introduces the propaganda techniques we considered in this shared task.
Section~\ref{sec:framework} describes the organization of the task, the corpus and the evaluation metrics. 
An overview of the participating systems is given in Section~\ref{sec:systems}, while Section~\ref{sec:results} discusses the evaluation results. 
Related work is described in Section~\ref{sec:pilot}. Finally, Section~\ref{sec:conclusions} draws some conclusions.

\section{Propaganda and its Techniques} 
\label{sec:propagandatechniques}

Propaganda comes in many forms, but it can be recognized by its persuasive function, sizable target audience, the representation of a specific group's agenda, and the use of faulty reasoning and/or emotional appeals~\cite{Miller}. 
The term \emph{propaganda} was coined in the 17th century, and initially referred to the propagation of the Catholic faith in the New World~\cite[p. 2]{Jowett:12}. It soon took a pejorative connotation, as its meaning was extended to also mean opposition to Protestantism. In more recent times, the Institute for Propaganda Analysis~\cite{InstituteforPropagandaAnalysis1938} proposed the following definition:
\begin{center}
  \parbox{39em}{\textbf{Propaganda.} \emph{Expression of opinion or action by individuals or groups deliberately designed to influence opinions or actions of other individuals or groups with reference to predetermined ends}.}%
\end{center}

\noindent
Recently, \newcite{Bolsover2017} dug deeper into this definition identifying its two key elements:
(\emph{i})~trying to influence opinion, and (\emph{ii})~doing so on purpose.

Propaganda is a broad concept, which runs short for the aim of annotating specific propaganda fragments. 
Yet, influencing opinions is achieved through a series of rhetorical and psychological techniques, and in the present task, we focus on identifying the use of such techniques in text.
Whereas the definition of propaganda is widely accepted in the literature, 
the set of propaganda techniques considered, and to some extent their definition, differ between different scholars~\cite{Torok2015}. For instance, 
\newcite{Miller} considers seven propaganda techniques, whereas~\newcite{Weston2000} lists at least 24 techniques, and the Wikipedia article on the topic includes 67.\footnote{\url{https://en.wikipedia.org/wiki/Propaganda_techniques}; last visit February 2019.} 
Below, we describe the propaganda techniques we consider in the task: a curated list of fourteen techniques derived from the aforementioned studies. We only include techniques that can be found in journalistic articles and can be judged intrinsically, without the need to retrieve supporting information from external resources. For example, we do not include techniques such as \emph{card stacking}~\cite[p.~237]{Jowett2012a}, since it would require comparing multiple sources. 
Note that our list of techniques was initially longer than fourteen, but we decided, after the annotation phase, to merge similar techniques with very low frequency in the corpus. A more detailed list with definitions and examples is available online,\footnote{\url{http://propaganda.qcri.org/annotations/definitions.html}}
and examples are shown in Table~\ref{tab:instances}.

\begin{table*}[tbh]
\renewcommand{\arraystretch}{1.25}%
\small
\centering
\begin{tabular}{@{}p{0.03\textwidth}@{}p{0.23\textwidth}@{}p{0.72\textwidth}}
\toprule
\bf \hspace*{1ex}\#	& \bf Technique						& \bf Snippet \\
\midrule
\hspace*{1ex}1	& {Loaded language}				&  \textbf{Outrage} as Donald Trump suggests injecting disinfectant to kill virus.\\
\hspace*{1ex}2	& {Name calling, labeling}			&  
WHO: Coronavirus emergency is '\textbf{Public Enemy Number 1}'\\
\hspace*{1ex}3	& {Repetition}					&  	I still have a \textbf{dream}. It is a \textbf{dream} deeply rooted in the American \textbf{dream}. I have a \textbf{dream} that one day \ldots\\
\hspace*{1ex}4	& {Exaggeration, minimization}		&  
Coronavirus \textbf{`risk to the American people remains very low'}, Trump said.\\
\hspace*{1ex}5	& {Doubt}						&  \textbf{Can the same be said for the 
Obama Administration}?	\\
\hspace*{1ex}6	& {Appeal to fear/prejudice}		&  \textbf{A dark, impenetrable 
and ``irreversible'' winter of persecution of the faithful by their own shepherds 
will fall}.	\\
\hspace*{1ex}7	& {Flag-waving}					&  Mueller attempts \textbf{to 
stop the will of We the People}!!! It's time to jail Mueller. 	\\
\hspace*{1ex}8	& {Causal oversimplification}		&  \textbf{If France had not have declared war on Germany then World War II would have never happened.} \\
\hspace*{1ex}9	& {Slogans}					&  \textbf{``BUILD THE WALL!''} 
Trump tweeted.\\
10			& {Appeal to authority}			&  \textbf{Monsignor 
Jean-François Lantheaume, who served as first Counsellor of the Nunciature in 
Washington, confirmed that ``Viganò said the truth. That's all.''}	\\
11			& {Black-and-white fallacy}		&  Francis said these words: 
``\textbf{Everyone is guilty for the good he could have done and did not do 
\ldots If we do not oppose evil, we tacitly feed it}.'' 	\\
12			& {Thought-terminating clich\'{e}}	&  \textbf{I do not really see 
any problems there.} Marx is the President.	\\
13			& {Whataboutism}				&  President Trump ---\textbf{who 
himself avoided national military service} in the 1960's--- keeps beating the 
war drums over North Korea.	\\ 
			& {Straw man}					&  ``Take it seriously, but with 
a large grain of salt.'' \textbf{Which is just Allen's more nuanced way of 
saying: ``Don't believe it}.''	\\
			& {Red herring}				&  \textbf{``You may claim that the death penalty is an ineffective deterrent against crime -- but what about the victims of crime? How do you think surviving family members feel when they see the man who murdered their son kept in prison at their expense? Is it right that they should pay for their son's murderer to be fed and housed?''} \\
14			& {Bandwagon}				&  He tweeted, ``\textbf{EU no 
longer considers \#Hamas a terrorist group. Time for US to do same.''} 	\\
\phantom{14}	& {Reductio ad hitlerum}			& ``Vichy journalism,'' a term which 
now fits so much of the mainstream media. \textbf{It collaborates in the same 
way that the Vichy government in France collaborated with the Nazis.}	\\
\bottomrule
\end{tabular}
\caption{The 14 propaganda techniques with examples, where the propaganda span is shown in bold. \label{tab:instances}}
\end{table*}

\paragraph{1. Loaded language.}
Using specific words and phrases with strong emotional implications (either positive or negative) to influence an audience~\cite[p.~6]{Weston2000}.
\vspace{-6pt}
\paragraph{2. Name calling or labeling.}
Labeling the object of the propaganda campaign as either something the target audience fears, hates, finds undesirable or loves, praises~\cite{Miller}.
\vspace{-6pt}
\paragraph{3. Repetition.} 
Repeating the same message over and over again, so that the audience will eventually accept it~\cite{Torok2015,Miller}.
\vspace{-6pt}
\paragraph{4. Exaggeration or minimization.}
Either representing something in an excessive manner: making things larger, better, worse or making something seem less important or smaller than it actually is \cite[pag. 303]{Jowett2012a}.
\vspace{-6pt}
\paragraph{5. Doubt.}
Questioning the credibility of someone or something.
\vspace{-6pt}
\paragraph{6. Appeal to fear/prejudice.} 
Seeking to build support for an idea by instilling anxiety and/or panic in the population towards an alternative, possibly based on preconceived judgments.
\vspace{-6pt}
\paragraph{7. Flag-waving.} 
Playing on strong national feeling (or with respect to any group, e.g., race, gender, political preference) to justify or promote an action or idea \cite{Hobbs2008}.
\vspace{-6pt}
\paragraph{8. Causal oversimplification.} Assuming a single cause or reason when there are multiple causes behind an issue. 
We include in the definition also \emph{scapegoating}, i.e.\ the transfer of the blame to one person or group of people without investigating the complexities of an issue.
\vspace{-6pt}
\paragraph{9. Slogans.}
A brief and striking phrase that may include labeling and stereotyping. Slogans tend to act as emotional appeals~\cite{As2015}.
\vspace{-6pt}
\paragraph{10. Appeal to authority.}
Stating that a claim is true simply because a valid authority or expert on the issue supports it, without any other supporting evidence~\cite{Goodwin2011}. We include in this technique the special case in which the reference is not an authority or an expert, although it is referred to as \emph{testimonial} in the literature~\cite[pag. 237]{Jowett2012a}.
\vspace{-6pt}
\paragraph{11. Black-and-white fallacy, dictatorship.}
Presenting two alternative options as the only possibilities, when in fact more possibilities exist \cite{Torok2015}. \emph{Dictatorship} is an extreme case: telling the audience exactly what actions to take, eliminating any other possible choice.
\vspace{-6pt}
\paragraph{12. Thought-terminating clich\'e.}
Words or phrases that discourage critical thought and meaningful discussion on a topic. They are typically short, generic sentences that offer seemingly simple answers to complex questions or that distract attention away from other lines of thought~\cite[p.~78]{Hunter2015}.
\vspace{-6pt}
\paragraph{13. Whataboutism, straw man, red herring.}
Here we merge together three techniques, which are relatively rare taken individually:
(\emph{i})~\emph{Whataboutism:} Discredit an opponent's position by charging them with hypocrisy without directly disproving their argument~\cite{Richter2017}.
(\emph{ii})~\emph{Straw man:} When an opponent's proposition is substituted with a similar one which is then refuted in place of the original~\cite{Walton1996}. 
\newcite[p.~78]{Weston2000} specifies the characteristics of the substituted proposition: ``caricaturing an opposing view so that it is easy to refute''.  
(\emph{iii})~\emph{Red herring:}
Introducing irrelevant material to the issue being discussed, so that everyone's attention is diverted away from the points made~\cite[p.~78]
{Weston2000}.
\paragraph{14. Bandwagon, reductio ad hitlerum.} 
Here we merge together two techniques, which are relatively rare taken individually:
(\emph{i})~\emph{Bandwagon.}
Attempting to persuade the target audience to join in and take the course of action because ``everyone else is taking the same action''~\cite{Hobbs2008}.
(\emph{ii})~\emph{Reductio ad hitlerum:} Persuading an audience to disapprove an action or idea by suggesting that it is popular with groups hated in contempt by the target audience. It can refer to any person or concept with a negative connotation~\cite{Aper2009}.
\medskip

We provided the definitions, together with some examples and an annotation schema, to professional annotators, so that they can manually annotate news articles. 
The annotators worked with an earlier version of the annotation schema, which contained 18 techniques \cite{EMNLP19DaSanMartino}. As some of these techniques were quite rare, which could cause data sparseness issues for the participating systems, for the purpose of the present SemEval-2020 task 11, we decided not to distinguish each of the four rarest techniques. In particular, we merged \emph{Red herring} and \emph{Straw man} with \emph{Whataboutism} (under technique 13), since all three techniques are trying to divert the attention to an irrelevant topic and away from the actual argument. We further merged \emph{Bandwagon} with \emph{Reductio ad hitlerum} (under technique 14), since they both try to approve/disapprove an action or idea by pointing to what is popular/unpopular. Finally, we dropped one rare technique, which we could not easily merge with other techniques: \emph{Obfuscation, Intentional vagueness, Confusion}. As a result, we reduced the original 18 techniques to 14.

\section{Evaluation Framework} 
\label{sec:framework}

The SemEval 2020 Task 11 evaluation framework consists of the \corpus corpus and evaluation measures for both the span identification and the technique classification subtasks. 
We describe the organization of the task in Section~\ref{sec:taskstructure}; here,  we focus on the dataset, the evaluation measure, and the organization setup. 

\subsection{The \corpus Corpus}

In order to build the \corpus corpus, we retrieved a sample of news articles 
from the period starting in mid-2017 and ending in early 2019. 
We selected 13 propaganda and 36 non-propaganda news media outlets, as labeled by Media Bias/Fact Check,\footnote{An initiative where professional journalists profile news outlets; 
\url{https://mediabiasfactcheck.com}.} 
and we retrieved articles from these sources. 
We deduplicated the articles on the basis of word 
$n$-grams matching~\cite{Barron:2009} and we discarded faulty entries (e.g.,~empty entries 
from blocking websites).

\begin{figure}
    \centering
    \includegraphics[scale=1.0]{img/text-and-annotation}
    \caption{Example of a plain-text article (left) and its annotation 
(right). The \emph{Start} and \emph{End} columns are the indices representing the character 
span of the spotted technique.}
    \label{fig:exampleArticle}
\end{figure}

The annotation job consisted of both spotting a propaganda snippet and, at the same time, labeling it with a specific propaganda technique. \pn{The annotation guidelines are shown in the appendix; they are also available online.\footnote{\url{https://propaganda.qcri.org/annotations/}}}
We ran the annotation in two phases:
\Ni two annotators label an article independently and
\Nii the same two annotators gather together with a \textit{consolidator} to discuss dubious instances (e.g.,~spotted only by one annotator, boundary discrepancies, label mismatch, etc.).
This protocol was designed after a pilot annotation stage, in which a relatively large number of snippets had been spotted by one annotator only. The annotation team consisted of six professional annotators from A Data Pro,\footnote{\url{https://www.aiidatapro.com}} 
trained to spot and label the propaganda snippets from free text. The job was carried out on an instance of the Anafora annotation platform~\cite{N13-3004}, 
which we tailored for our propaganda annotation task. 
Figure~\ref{fig:exampleArticle} shows an example of an article and its annotations.

We evaluated the annotation process in terms of $\gamma$ 
agreement~\cite{Mathet2015} between each of the annotators and the final gold labels. 
The $\gamma$ agreement on the annotated articles is on average $0.6$; \pn{see \cite{EMNLP19DaSanMartino} for a more detailed discussion of inter-annotator agreement}. 
The training and the development part of the PTC-SemEval20 corpus are the same as the training and the testing datasets described in \cite{EMNLP19DaSanMartino}.
\pn{The test part of the PTC-SemEval20 corpus consists of 90 additional articles selected  from the same sources as for training and development.}
For the test articles, we further extended the annotation process by adding one extra consolidation step: we revisited all the articles in that partition and we performed the necessary adjustments to the spans and to the labels as necessary, after a thorough discussion and convergence among at least three experts who were not involved in the initial annotations. 

\begin{figure}[t]
\centering
\includegraphics[width=1.08\textwidth]{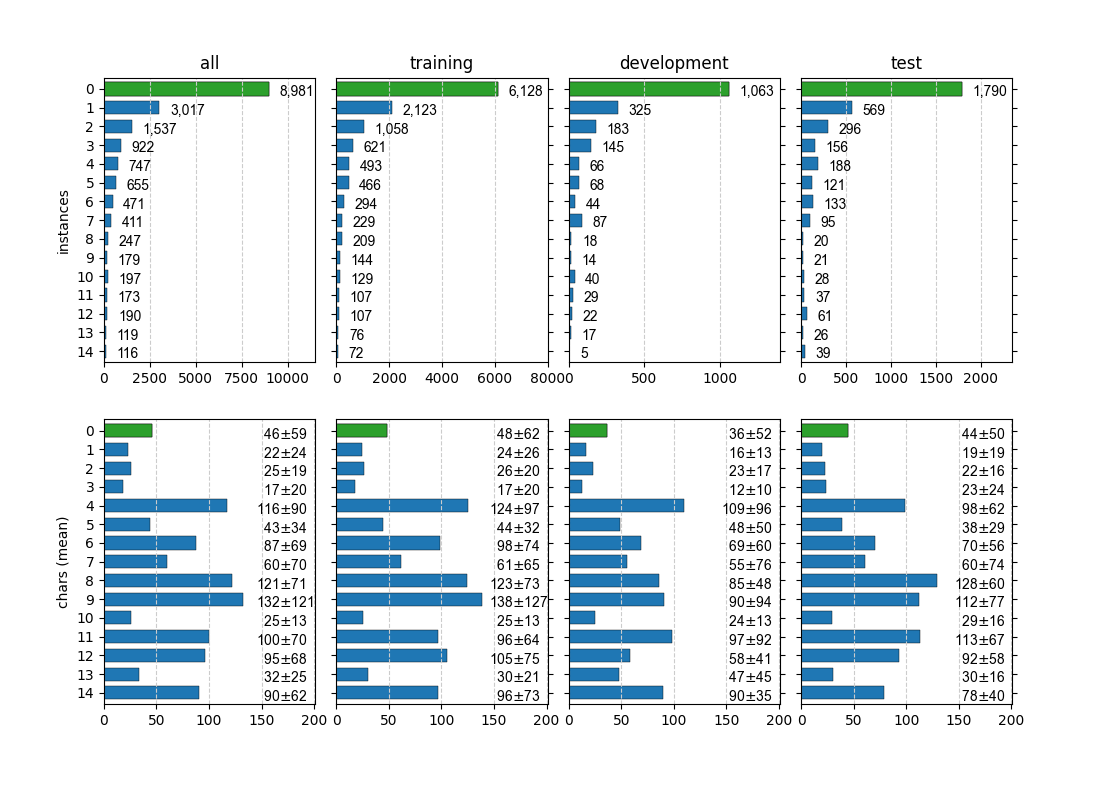}
\footnotesize

\vspace{-10mm}
\begin{tabular}{rlrlrlrl}
0 & Overall			& 5 & Doubt			& 10	& Appeal 
to authority	\\
1 & Loaded language		& 6 & Appeal to fear/prejudice	& 11	& 
Black-and-white fallacy, dictatorship	\\
2 & Name calling or labeling	& 7 & Flag-waving		& 12	& 
Thought-terminating clich\'e	\\
3 & Repetition			& 8 & Causal oversimplification & 13	& 
Whataboutism, straw man, red herring	\\
4 & Exaggeration or minimization& 9 & Slogans			& 14	& 
Bandwagon, reductio ad hitlerum	\\
\end{tabular}
  \caption{Statistics on the propaganda snippets in the different 
partitions of the \corpus corpus. Top: number of instances per class. 
Bottom: mean snippet length per class.}
  \label{fig:statscorpus}
\end{figure}

Table~\ref{tab:statscorpus} shows some corpus statistics. 
It is worth noting 
that a number of propaganda snippets of different classes overlap. Hence, the 
number of snippets for the span identification subtask is smaller (e.g.,
~1,405
for the span identification subtask vs. 1,790
for the technique classification subtask on 
the test set). The full collection of 536 articles contains 8,981 
propaganda 
text snippets, belonging to one of fourteen possible classes. 
Figure~\ref{fig:statscorpus} zooms into such snippets and shows the number 
of instances and the mean lengths for all classes. By a large margin, the most 
common propaganda technique in news articles is \emph{Loaded Language}, which is about twice as frequent as the second most frequent technique: \emph{Name Calling or 
Labeling}. Whereas these two techniques are among the ones that are expressed in 
the shortest spans, other techniques such as \emph{Exaggeration}, \emph{Causal 
Oversimplification}, and \emph{Slogans} tend to be the longest. 

\begin{table}
\centering
\begin{tabular}{l|cccc}
\toprule
\bf partition& \bf articles	& \multicolumn{2}{c}{\bf average lengths}& \bf propaganda	\\
		    & 	        & \bf chars         & \bf tokens	& \bf snippets\\
\midrule
training	& 371	    & 5,681$\pm$5,425	& 927$\pm$899	& 6,128\\
development	& \,\,75	& 4,700$\pm$2,904	& 770$\pm$473	& 1,063	\\
test		& \,\,90	& 4,518$\pm$2,602	& 744$\pm$433	& 1,790	\\
\rowcolor{Gray}
all		&  536	& 5,348$\pm$4,789	& 875$\pm$793	& 8,981\\
\bottomrule
\end{tabular}
\caption{Statistics about the \corpus corpus including the number of articles, average lengths in terms of characters and tokens, 
and number of propaganda snippets.}
\label{tab:statscorpus}
\end{table}

\subsection{The Evaluation Measures}

\newcommand{\character}{\ensuremath{s}\xspace}
\newcommand{\article}{\ensuremath{d}\xspace}
\newcommand{\dataset}{\ensuremath{D}\xspace}
\newcommand{\anyspan}{\ensuremath{s}\xspace}
\newcommand{\gold}{\ensuremath{t\xspace}}
\newcommand{\Gold}{\ensuremath{T}\xspace}
\newcommand{\pred}{\ensuremath{s}\xspace}
\newcommand{\Pred}{\ensuremath{S}\xspace}
\newcommand{\precision}{\ensuremath{P}\xspace}
\newcommand{\recall}{\ensuremath{R}\xspace}
\newcommand{\fscore}{\ensuremath{F_1}\xspace}

\paragraph{Subtask SI} 
Evaluating subtask SI requires to match text spans. 
Our SI evaluation function gives credit to partial matches between gold and predicted spans. 

\begin{figure}
    \centering
    \includegraphics[scale=1.0]{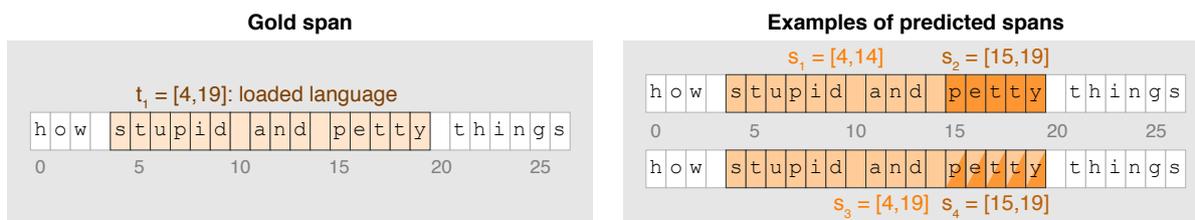}
    \caption{Example of equivalent annotations for the Span Identification subtask.}
     \label{fig:si_example}
\end{figure}

Let $\article$ be a news article in a set $\dataset$. 
A gold span $t$ is a sequence of contiguous indices of the characters composing a text fragment $\gold\subseteq d$. 
For example, in Figure~\ref{fig:si_example} (top-left) the gold fragment \emph{``stupid and petty''} is represented by the set of indices $t_1=[4, 19]$. 
We denote with $\Gold_d=\{\gold_1, \ldots, \gold_ n\}$ the set of all gold spans for an article $d$ and with $\Gold=\{\Gold_d\}_d$ the set of all gold annotated spans in $\dataset$. 
Similarly, we define $\Pred_d = \{ \pred_1, \ldots, \pred_m\}$ and $\Pred$ to be the set of predicted spans for an article $d$ and a dataset $\dataset$, respectively. 
We  compute precision $\precision$ and recall $\recall$ by adapting the formulas in~\cite{Potthast2010a}:
%
\begin{equation}
\footnotesize
 \precision(\Pred, \Gold) \quad=\quad \frac{1}{|\Pred|} \cdot \sum_{\article \in \dataset}\sum_{\pred\in \Pred_\article, \gold \in \Gold_\article} \frac{|(\pred \cap \gold)|}{|\gold|},
    \label{eq:spanPrecision}
\end{equation}
\begin{equation}
\footnotesize
 \recall(\Pred, \Gold) \quad=\quad \frac{1}{|\Gold|} \cdot \sum_{\article \in \dataset}\sum_{\pred\in \Pred_\article, \gold \in \Gold_\article}\!\!\! \frac{|(\pred \cap \gold)|}{|\pred|}.  
\label{eq:spanRecall}
\end{equation}
%
We define \mbox{Eq.~(\ref{eq:spanPrecision})} to be zero when $|\Pred|=0$ and \mbox{Eq.~(\ref{eq:spanRecall})} to be zero when $|\Gold|=0$. 
Notice that the predicted spans may overlap, e.g.,~spans $s_3$ and $s_4$ in Figure~\ref{fig:si_example}. Therefore, in order for Eq.~\ref{eq:spanPrecision} and Eq.~\ref{eq:spanRecall} to get values lower than or equal to 1, all overlapping annotations, independently of their techniques, are merged first. 
For example, $s_3$ and $s_4$ are merged into one single annotation, corresponding to $s_4$. 
Finally, the evaluation measure for \mbox{subtask SI} is the $\fscore$ score, defined as the  harmonic mean between $\precision(\Pred, \Gold)$ and $ \recall(\Pred, \Gold)$: 
\begin{equation}
\footnotesize
    \fscore(\Pred, \Gold)=2 \cdot \frac{\precision(\Pred, \Gold) \cdot \recall(\Pred, \Gold)}{\precision(\Pred, \Gold) + \recall(\Pred, \Gold)}.
    \label{eq:f}
\end{equation}

\paragraph{Subtask TC} 
Given a propaganda snippet in an article, subtask TC asks to identify the technique in it. 
Since there are identical spans annotated with different techniques (around 1.8\% of the total annotations), formally this is a multi-label multi-class classification problem. 
However, we decided to consider the problem as a single-label multi-class one, by performing the following adjustments: 
\Ni whenever a span is associated with multiple techniques, the input file will have multiple copies of such fragments and
\Nii the evaluation function ensures that the best match between the predictions and the gold labels for identical spans is used for the evaluation. In other words, the evaluation score is not affected by the order in which the predictions for identical spans are submitted. 

The evaluation measure for subtask TC is micro-average F$_1$. Note that as we have converted this to a single-label task, micro-average F$_1$ is equivalent to Accuracy (as well as to Precision and to Recall).

\subsection{Task Organization}\label{sec:taskstructure}

The shared task was divided in two phases:

\paragraph{Phase 1.} Only training and development data were made available, and 
no gold labels were provided for the latter. The participants competed against 
each other to achieve the best performance on the development set. A live 
leaderboard was made available to keep track of all submissions.

\paragraph{Phase 2.} The test set was released and the participants were given 
just a few days to submit their final predictions. 
The release of the test set was done task-by-task, since giving access to the input files for the TC subtask would have disclosed the gold spans for the SI subtask.

\medskip
In phase~1, the participants could make an unlimited number of submissions on the development set, and they could see the outcomes in their private space. The best 
team score, regardless of the submission time, was also shown in a public 
leaderboard. As a result, not only could the participants observe 
the impact of their systems modifications, but they could also compare against the results by other participating teams. 
In phase~2, the participants could again submit multiple runs, but they did not get 
any feedback on their performance. Only the last submission of each team was 
considered official and used for the final ranking.

In phase 1, a total of 47 teams made submissions on the development set for the SI subtask, and 46 teams submitted for the TC subtask. In phase 2, the number of teams who made official submissions on the test set for subtasks SI and TC was 35 and 31, respectively: this is a total of 66 submissions for the two subtasks, which were made by 44 different teams.

Note that we left the submission system open for submissions on the development set (phase 1) after the competition was over. The up-to-date leaderboards can be found on the website of the competition.\footnote{\url{http://propaganda.qcri.org/semeval2020-task11/leaderboard.php}}



\section{Participating Systems} \label{sec:systems}

In this section, we focus on a general description of the systems participating 
on both the SI and the TC subtasks. We pay special attention to the most 
successful approaches. The subindex on the right of each team represents their 
official rank in the subtasks. Appendix~\ref{app:full_description} includes brief descriptions of all systems. 

\subsection{Span Identification Subtask}\label{sec:participantsi}

Table~\ref{tab:overview_si} shows a quick overview of the systems that took part in
the SI subtask.%
\footnote{Tables~\ref{tab:overview_si} and~\ref{tab:overview_tc} only include the systems for which a description paper has been submitted.}
All systems in the top-10 positions relied on some kind of Transformer, used in combination with an LSTM or a CRF. In most cases, the Transformer-generated representations were complemented by engineered features, 
such as named entities and presence of sentiment and subjectivity clues.

Team \textbf{Hitachi}\ranksi{1} achieved the top performance in this subtask~\cite{SemEval20-11-Morio}. 
They used a BIO encoding, which is typical for related segmentation and labeling 
tasks (e.g., named entity recognition). They relied upon a complex heterogeneous 
multi-layer neural network, trained end-to-end. The network uses pre-trained 
language models, which 
generate a representation for each input token. To this are added both part-of-speech 
(PoS) and named entity (NE) embeddings. As a result, there are three 
representations for each token, which are concatenated and used as an input to 
bi-LSTMs. At this moment, the network branches, as it is trained with three 
objectives: (\emph{i})~the main BIO tag prediction objective and two auxiliary 
ones, namely (\emph{ii})~token-level technique classification, and 
(\emph{iii})~sentence-level classification. There is one Bi-LSTM for objectives 
(\emph{i})~and (\emph{ii}), and there is another Bi-LSTM for objective 
(\emph{iii}). For the former, they use an additional CRF layer, which helps 
improve the consistency of the output. A number of architectures are trained 
independently ---using BERT, GPT-2, XLNet, XLM, RoBERTa, or XLM-RoBERTa---, and 
the resulting models are combined in ensembles.

Team \textbf{ApplicaAI}\ranksi{2}~\cite{SemEval20-11-Jurkiewicz} based its success on 
self-supervision using the RoBERTa model. They used a RoBERTa-CRF architecture 
trained on the provided data and used it to iteratively produce silver data by 
predicting on 500k sentences and retraining the model with both gold and silver 
data. The final classifier was an ensemble of models trained on the original 
corpus, re-weighting, and a model trained also on silver data. ApplicaAI was not 
the only team that obtained a performance boost by considering additional data. 
Team \textbf{UPB}\ranksi{5}~\cite{SemEval20-11-Paraschiv} decided not to stick to the 
pre-trained models from BERT--base alone and used masked language modeling to 
domain-adapt it using 9M articles containing fake, suspicious, and hyperpartisan news 
articles.
Team \textbf{DoNotDistribute}\ranksi{22}~\cite{SemEval20-11-Kranzlein} also opted for generating 
silver data, but with a different strategy. They report a 5\% performance boost when adding 3k new silver training instances. To produce them, 
they used a library to create near-paraphrases of the propaganda snippets by 
randomly substituting certain PoS words.
Team \textbf{SkoltechNLP}\ranksi{25}~\cite{SemEval20-11-Dementieva} performed data 
augmentation based on distributional semantics.
Finally, team \textbf{WMD}\ranksi{33}~\cite{SemEval20-11-Daval-Frerot}  applied multiple strategies 
to augment the data such as back translation, synonym replacement and TF.IDF replacement 
(replace unimportant words, based on TF.IDF score, by other 
unimportant words). 

Closing the top-three submissions, Team \textbf{aschern}\ranksi{3}~\cite{SemEval20-11-Chernyavskiy} 
fine-tuned an ensemble of two differently intialized RoBERTa models, each with 
an attached CRF for sequence labeling and simple span character boundary 
post-processing. 

There have been several other promising strategies. Team 
\textbf{LTIatCMU}\ranksi{4}~\cite{SemEval20-11-Khosla} used a multi-granular BERT BiLSTM model
that employs additional syntactic and semantic features at the word, sentence and document level, including PoS, named entities, sentiment, and subjectivity.   It was trained jointly for token and sentence propaganda classification, with class balancing. They further fine-tuned BERT on persuasive language using 10,000 articles from propaganda websites, 
which turned out to be important in their experiments. 
Team \textbf{PsuedoProp}\ranksi{14}~\cite{SemEval20-11-Chauhan} opted for building a 
preliminary sentence-level classifier using an ensemble of XLNet and RoBERTa, 
before it fine-tuned a BERT-based CRF sequence tagger to identify the exact spans.
Team \textbf{BPGC}\ranksi{21}~\cite{SemEval20-11-Patil} went beyond these multigranularity 
approaches. Information both at the full article and at the sentence level was 
considered when classifying each word as propaganda or not, by means of 
computing and concatenating vectorial representations for the three inputs.

A large number of teams decided to rely heavily on engineered features. For instance, 
Team \textbf{CyberWallE}\ranksi{8}~\cite{SemEval20-11-Blaschke} used features modeling sentiment, rhetorical structure, and POS tags, while team \textbf{UTMN}\ranksi{23} injected the sentiment intensity from VADER and was among the only teams not relying on deep-learning architectures in order to produce a computationally-affordable model.

\newcolumntype{g}{>{\columncolor{Gray}}c}
\begin{table}[t]
\centering
\setlength{\tabcolsep}{1.7pt}    
\footnotesize
\begin{tabular}{rl|ggggggggg |
		  gggggggg |
		  ggggggggggg | 
		  ggg
		  }
\toprule		 
\multicolumn{2}{l}{\bf Rank. Team} & \multicolumn{9}{c}{\bf Transformers}	& \multicolumn{8}{c}{\bf Learning Models}	& \multicolumn{11}{c}{\bf Representations} & \multicolumn{3}{c}{\bf Misc}	\\
\rowcolor{white}
	&   &
\rotatebox{90}{\textbf{BERT }} &
\rotatebox{90}{\textbf{RoBERTa }} &
\rotatebox{90}{\textbf{XLNet }} &
\rotatebox{90}{\textbf{XLM}} &
\rotatebox{90}{\textbf{XLM RoBERTa}} &
\rotatebox{90}{\textbf{ALBERT}} &
\rotatebox{90}{\textbf{GPT-2 }} &
\rotatebox{90}{\textbf{SpanBERT }} &
\rotatebox{90}{\textbf{LaserTagger}} &
\rotatebox{90}{\textbf{LSTM }} &
\rotatebox{90}{\textbf{CNN }} &
\rotatebox{90}{\textbf{SVM }} &
\rotatebox{90}{\textbf{Na\"ive Bayes }} &
\rotatebox{90}{\textbf{Boosting}} &
\rotatebox{90}{\textbf{Log regressor }} &
\rotatebox{90}{\textbf{Random forest}} &
\rotatebox{90}{\textbf{CRF }} &  
\rotatebox{90}{\textbf{Embeddings }} &
\rotatebox{90}{\textbf{ELMo }} &
\rotatebox{90}{\textbf{NEs }} &
\rotatebox{90}{\textbf{Words/$n$-grams}} &
\rotatebox{90}{\textbf{Chars/$n$-grams}} &
\rotatebox{90}{\textbf{PoS }} &
\rotatebox{90}{\textbf{Trees }} &
\rotatebox{90}{\textbf{Sentiment }} &
\rotatebox{90}{\textbf{Subjectivity }} &
\rotatebox{90}{\textbf{Rhetorics }} &
\rotatebox{90}{\textbf{LIWC}} &

\rotatebox{90}{\textbf{Ensemble}}   &
\rotatebox{90}{\textbf{Data augmentation}}   &
\rotatebox{90}{\textbf{Post-processing}}
\\ \midrule
\rowcolor{white}
1.	& Hitachi	& \sq	& \sq	& \sq	& \sq	& \sq	&  	& \sq	&  	&  	&  	& \sq	&  	&  	&  	&  	&  	& \sq	& \sq	&  	& \sq	&  	&  	& \sq	&  	&  	&  	&  	&  	& \sq	&  	&  	\\
2.	& ApplicaAI	&  	& \sq	&  	&  	&  	&  	&  	&  	&  	&  	&  	&  	&  	&  	&  	&  	& \sq	&  	&  	&  	&  	&  	&  	&  	&  	&  	&  	&  	&  	& \sq	&  	\\
\rowcolor{white}
3.	& aschern	&  	& \sq	&  	&  	&  	&  	&  	&  	&  	&  	&  	&  	&  	&  	&  	&  	& \sq	& \sq	&  	&  	&  	& \sq	&  	&  	&  	&  	&  	&  	& \sq	&  	& \sq	\\
4.	& LTIatCMU	& \sq	&  	&  	&  	&  	&  	&  	&  	&  	& \sq	&  	&  	&  	&  	&  	&  	&  	& \sq	&  	& \sq	& \sq	&  	& \sq	& \sq	& \sq	& \sq	&  	& \sq	&  	&  	&  	\\
\rowcolor{white}
5.	& UPB	& \sq	&  	&  	&  	&  	&  	&  	&  	&  	& \cq	&  	&  	&  	&  	&  	&  	& \sq	& \cq	&  	&  	&  	&  	&  	&  	&  	&  	&  	&  	&  	& \sq	&  	\\
7.	& NoPropaganda	& \sq	&  	&  	&  	&  	&  	&  	&  	& \sq	& \cq	&  	&  	&  	&  	&  	&  	& \cq	& \sq	& \sq	&  	&  	& \sq	&  	&  	&  	&  	&  	&  	&  	&  	&  	\\
\rowcolor{white}
8.	& CyberWallE	& \sq	&  	&  	&  	&  	&  	&  	&  	&  	& \sq	&  	& \cq	&  	&  	&  	&  	&  	&  	&  	&  	&  	&  	& \sq	&  	& \sq	&  	& \sq	&  	&  	&  	&  	\\
9.	& Transformers	& \sq	& \sq	&  	&  	&  	&  	&  	&  	&  	& \sq	& \sq	&  	&  	&  	&  	&  	&  	&  	& \sq	&  	&  	&  	&  	&  	&  	&  	&  	&  	& \sq	&  	&  	\\
\rowcolor{white}
11.	& YNUtaoxin	& \sq	& \cq	& \cq	&  	&  	&  	&  	&  	&  	&  	&  	&  	&  	&  	&  	&  	&  	&  	&  	&  	&  	&  	&  	&  	&  	&  	&  	&  	&  	&  	&  	\\
13.	& newsSweeper	& \sq	& \cq	&  	&  	&  	&  	& \cq	& \cq	&  	&  	&  	&  	&  	&  	&  	&  	&  	&  	&  	&  	&  	&  	& \cq	&  	&  	&  	&  	&  	&  	&  	&  	\\
\rowcolor{white}
14.	& PsuedoProp	& \sq	& \sq	& \sq	&  	&  	& \cq	&  	&  	&  	&  	&  	&  	&  	&  	&  	&  	&  	&  	&  	&  	&  	&  	&  	&  	&  	&  	&  	&  	& \sq	&  	&  	\\
16.	& YNUHPCC	& \cq	&  	&  	&  	&  	&  	&  	&  	&  	& \sq	&  	&  	&  	&  	&  	&  	&  	& \sq	&  	&  	&  	&  	&  	&  	&  	&  	&  	&  	&  	&  	&  	\\
\rowcolor{white}
17.	& NLFIIT	& \cq	&  	&  	&  	&  	&  	&  	&  	&  	& \sq	&  	&  	&  	&  	&  	&  	& \cq	& \cq	& \sq	&  	&  	&  	&  	&  	&  	&  	&  	&  	& \cq	&  	&  	\\
20.	& TTUI	& \sq	& \sq	&  	&  	&  	&  	&  	&  	&  	&  	&  	&  	&  	&  	&  	&  	&  	&  	&  	&  	& \sq	&  	&  	&  	&  	&  	&  	&  	& \sq	&  	& \sq	\\
\rowcolor{white}
21.	& BPGC	& \cq	& \sq	& \cq	&  	&  	&  	& \cq	&  	&  	&  	&  	&  	&  	&  	&  	&  	&  	& \sq	&  	&  	&  	&  	&  	&  	&  	&  	&  	&  	&  	&  	&  	\\
22.	& DoNotDistribute	& \sq	&  	&  	&  	&  	&  	&  	&  	&  	& \cq	& 	&  	&  	&  	&  	&  	&  	&  	&  	& \sq	& \sq	&  	& \sq	&  	&  	&  	&  	&  	&  	& \sq	&  	\\
\rowcolor{white}
23.	& UTMN	&  	&  	&  	&  	&  	&  	&  	&  	&  	&  	&  	&  	&  	&  	& \sq	&  	&  	& \sq	&  	&  	&  	&  	&  	&  	& \sq	&  	&  	&  	&  	&  	&  	\\
25.	& syrapropa	&  	&  	&  	&  	&  	&  	&  	& \sq	&  	&  	&  	&  	&  	&  	&  	&  	&  	& \sq	&  	&  	& \sq	&  	&  	&  	&  	&  	&  	&  	&  	&  	&  	\\
\rowcolor{white}
26.	& SkoltechNLP	& \sq	&  	&  	&  	& 	&  	&  	&  	&  	& \cq	& \cq	&  	&  	&  	&  	&  	& \cq	&  	& \sq	&  	&  	& \sq	&  	&  	&  	&  	&  	&  	&  	&  	&  	\\
27.	& NTUAAILS	&  	&  	&  	&  	&  	&  	&  	&  	&  	& \sq	&  	&  	&  	&  	&  	&  	&  	&  	& \sq	&  	&  	&  	&  	&  	&  	&  	&  	&  	&  	&  	&  	\\
\rowcolor{white}
28.	& UAIC1860	&  	&  	&  	&  	&  	&  	&  	&  	&  	&  	&  	& \cq	& \cq	&  	& \cq	& \sq	&  	& \sq	&  	&  	& \sq	& \sq	&  	&  	&  	&  	&  	&  	&  	&  	&  	\\
31.	& 3218IR	&  	&  	&  	&  	&  	&  	&  	&  	&  	&  	& \sq	&  	&  	&  	&  	&  	&  	& \sq	&  	&  	&  	&  	&  	&  	&  	&  	&  	&  	&  	&  	&  	\\
\rowcolor{white}
33.	& WMD	& \sq	&  	&  	&  	&  	&  	&  	&  	&  	&  	&  	& \cq	&  	& \cq	&  	& \cq	&  	& \sq	&  	&  	&  	&  	&  	&  	&  	&  	&  	&  	& \sq	& \sq	& \sq	\\
-- & UNTLing	&  	&  	&  	&  	&  	&  	&  	&  	&  	&  	&  	&  	&  	&  	&  	&  	& \cq	& \cq	&  	&  	& \cq	&  	& \cq	& \cq	& \cq	&  	&  	&  	&  	&  	&  	\\
\bottomrule
\end{tabular}

\begin{tabular}{rl}
1.	& \cite{SemEval20-11-Morio}	\\	
2.	& \cite{SemEval20-11-Jurkiewicz}	\\	
3.	& \cite{SemEval20-11-Chernyavskiy}	\\	
4.	& \cite{SemEval20-11-Khosla}		\\	
5.	& \cite{SemEval20-11-Paraschiv}	\\	
7.	& \cite{SemEval20-11-Dimov}		\\	
8.	& \cite{SemEval20-11-Blaschke}	\\	
9.	& \cite{SemEval20-11-Verma}		\\	
\end{tabular}
\hspace{8mm}
\begin{tabular}{rl}
11.	& \cite{SemEval20-11-Tao}		\\	
13.	& \cite{SemEval20-11-Singh}	\\	
14.	& \cite{SemEval20-11-Chauhan}	\\	
16.	& \cite{SemEval20-11-Dao}	\\	
17.	& \cite{SemEval20-11-Martinkovic}	\\	
20.	& \cite{SemEval20-11-Kim}		\\	
21.	& \cite{SemEval20-11-Patil}		\\	
22.	& \cite{SemEval20-11-Kranzlein}	\\	
\end{tabular}
\hspace{8mm}
\begin{tabular}{rl}
23.	& \cite{SemEval20-11-Mikhalkova}	\\	
25.	& \cite{SemEval20-11-Li}	\\	
26.	& \cite{SemEval20-11-Dementieva}	\\	
27.	& \cite{SemEval20-11-Arsenos}	\\	
28.	& \cite{SemEval20-11-Ermurachi}	\\	
31.	& \cite{SemEval20-11-Dewantara}	\\	
33.	& \cite{SemEval20-11-Daval-Frerot}	\\	
--	& \cite{SemEval20-11-Krishnamurthy}	\\
\end{tabular}

\caption{Overview of the approaches to the span identification subtask. \sq$=$part 
of the official submission; \cq$=$considered in internal experiments. The 
references to the description papers appear at the bottom.}
\label{tab:overview_si}
\end{table}

\subsection{Technique Classification Subtask}

The same trends as for the snippet identification subtask can be observed in the 
approaches used for the technique classification subtask: practically all the 
top-performing approaches used representations produced by some kind of Transformer.

Team \textbf{ApplicaAI}\ranktc{1} achieved the top performance for this 
subtask~\cite{SemEval20-11-Jurkiewicz}. As in their approach to subtask SI, ApplicaAI produced 
silver data to train on. This 
time, they ran their high-performing SI model to spot new propaganda snippets 
on free text and applied their preliminary TC model to produce extra 
silver-labeled instances. Their final classifier consisted of an ensemble of 
models trained on the original corpus, re-weighting, and a model trained also on 
silver data. In all cases, the input to the classifiers consisted of 
propaganda snippets and their context. 

Team \textbf{aschern}\ranktc{2}~\cite{SemEval20-11-Chernyavskiy} based its success on a 
RoBERTa ensemble. They treated the task as sequence classification, using an average embedding of the surrounding tokens and the length of the span as 
contextual features. Via transfer learning, knowledge from both subtasks was 
incorporated. They further performed specific postprocessing to 
increase the consistency for the repetition technique spans and to avoid 
insertions of techniques in other techniques.

\begin{table}[t]
\centering
\setlength{\tabcolsep}{1.07pt}  
\footnotesize
\begin{tabular}{rl|gggggggggg |
		  ggggggggggg |
		  ggggggggggg | 
		  ggg
		  }
\toprule		 
\multicolumn{2}{l}{\bf Rank. Team}	& \multicolumn{10}{c}{\bf Transformers}	& \multicolumn{11}{c}{\bf Learning Models}	& \multicolumn{11}{c}{\bf Representations} & \multicolumn{3}{c}{\bf Misc}	\\
\rowcolor{white}
&	&
\rotatebox{90}{\textbf{BERT}} &
\rotatebox{90}{\textbf{R BERT }} &
\rotatebox{90}{\textbf{RoBERTa }} &
\rotatebox{90}{\textbf{XLNet }} &
\rotatebox{90}{\textbf{XLM}} &
\rotatebox{90}{\textbf{XLM RoBERTa}} &
\rotatebox{90}{\textbf{ALBERT}} &
\rotatebox{90}{\textbf{GPT-2 }} &
\rotatebox{90}{\textbf{SpanBERT }} &
\rotatebox{90}{\textbf{DistilBERT }} &
\rotatebox{90}{\textbf{LSTM }} &
\rotatebox{90}{\textbf{RNN }} &
\rotatebox{90}{\textbf{CNN }} &
\rotatebox{90}{\textbf{SVM }} &
\rotatebox{90}{\textbf{Na\"ive Bayes }} &
\rotatebox{90}{\textbf{Boosting }} &
\rotatebox{90}{\textbf{Log regressor }} &
\rotatebox{90}{\textbf{Random forest }} &
\rotatebox{90}{\textbf{Regression tree }} &
\rotatebox{90}{\textbf{CRF }} &  
\rotatebox{90}{\textbf{XGBoost }} &  
\rotatebox{90}{\textbf{Embeddings }} &
\rotatebox{90}{\textbf{ELMo }} &
\rotatebox{90}{\textbf{NEs }} &
\rotatebox{90}{\textbf{Words/$n$-grams}} &
\rotatebox{90}{\textbf{Chars/$n$-grams}} &
\rotatebox{90}{\textbf{PoS }} &
\rotatebox{90}{\textbf{Sentiment }} &
\rotatebox{90}{\textbf{Rhetorics }} &
\rotatebox{90}{\textbf{Lexicons}} &
\rotatebox{90}{\textbf{String matching}} &
\rotatebox{90}{\textbf{Topics }} &
\rotatebox{90}{\textbf{Ensemble}} &
\rotatebox{90}{\textbf{Data augmentation}} &
\rotatebox{90}{\textbf{Post-processing}}
\\ \midrule
\rowcolor{white}
1.	& ApplicaAI	&  	&  	& \sq	&  	&  	&  	&  	&  	&  	&  	&  	&  	&  	&  	&  	&  	&  	&  	&  	& \sq	&  	&  	&  	&  	&  	&  	&  	&  	&  	&  	&  	&  	& \sq	& \sq	&  	\\
2.	& aschern	&  	&  	& \sq	&  	&  	&  	&  	&  	&  	&  	&  	&  	&  	&  	&  	&  	&  	&  	&  	&  	&  	& \sq	&  	&  	&  	&  	&  	&  	&  	&  	&  	&  	& \sq	&  	& \sq	\\
\rowcolor{white}
3.	& Hitachi	& \sq	&  	& \sq	& \sq	& \sq	& \sq	& \sq	&  	&  	&  	&  	&  	& \sq	&  	&  	&  	&  	&  	&  	&  	&  	& \sq	&  	& \sq	&  	&  	& \sq	&  	&  	&  	&  	&  	& \sq	&  	&  	\\
4.	& Solomon	&  	&  	& \sq	&  	&  	&  	&  	&  	&  	&  	&  	&  	&  	& \cq	&  	&  	&  	&  	&  	&  	&  	& \sq	&  	&  	&  	&  	&  	&  	&  	&  	& \sq	&  	& \sq	&  	&  	\\
\rowcolor{white}
5.	& newsSweeper	& \cq	&  	& \sq	&  	&  	&  	&  	& \cq	& \cq	&  	&  	&  	&  	&  	&  	&  	&  	&  	&  	&  	&  	&  	&  	&  	&  	&  	& \cq	&  	&  	&  	&  	&  	&  	&  	&  	\\
6.	& NoPropaganda	& \sq	& \sq	&  	&  	&  	&  	&  	&  	&  	&  	&  	&  	&  	&  	&  	&  	&  	&  	&  	&  	&  	&  	&  	&  	&  	&  	&  	&  	&  	&  	&  	&  	&  	&  	&  	\\
\rowcolor{white}
7.	& Inno	& \cq	&  	& \sq	& \cq	&  	&  	&  	&  	&  	& \cq	& \cq	&  	&  	&  	&  	&  	& \cq	&  	&  	&  	&  	&  	& \cq	&  	&  	&  	&  	&  	&  	&  	&  	&  	&  	&  	&  	\\
8.	& CyberWallE	& \sq	&  	&  	&  	&  	&  	&  	&  	&  	&  	&  	&  	&  	&  	&  	&  	&  	&  	&  	&  	&  	&  	&  	& \sq	&  	&  	&  	&  	& \sq	&  	&  	&  	&  	&  	& \sq	\\
\rowcolor{white}
10.	& Duth	& \sq	&  	&  	&  	&  	&  	&  	&  	&  	&  	&  	&  	&  	&  	&  	&  	&  	&  	&  	&  	&  	&  	&  	& \sq	&  	&  	&  	&  	&  	&  	&  	&  	&  	&  	&  	\\
11.	& DiSaster	& \sq	&  	&  	&  	&  	&  	&  	&  	&  	&  	&  	& \cq	&  	&  	&  	&  	& \sq	&  	&  	&  	&  	&  	&  	&  	& \sq	&  	&  	&  	&  	&  	&  	&  	& \sq	&  	&  	\\
\rowcolor{white}
13.	& SocCogCom	& \sq	&  	&  	&  	&  	&  	& \cq	&  	&  	&  	&  	&  	&  	&  	&  	&  	&  	&  	& \cq	&  	&  	&  	&  	&  	&  	&  	&  	& \sq	&  	& \cq	&  	&  	&  	&  	&  	\\
14.	& TTUI	& \sq	&  	& \sq	&  	&  	&  	&  	&  	&  	&  	&  	&  	&  	&  	&  	&  	&  	&  	&  	&  	&  	& 	&  	&  	& \sq	&  	&  	&  	&  	&  	&  	&  	& \sq	&  	&  	\\
\rowcolor{white}
15.	& JUST	& \sq	&  	&  	&  	&  	&  	&  	&  	&  	&  	&  	&  	&  	&  	&  	&  	&  	&  	&  	&  	&  	& \sq	&  	&  	&  	&  	&  	&  	&  	&  	&  	&  	&  	&  	&  	\\
16.	& NLFIIT	& \cq	&  	&  	&  	&  	&  	&  	&  	&  	&  	& \sq	&  	&  	&  	&  	&  	&  	&  	&  	& \cq	&  	& \cq	& \sq	&  	&  	&  	&  	&  	&  	&  	&  	&  	& \cq	&  	&  	\\
\rowcolor{white}
17.	& UMSIForeseer	& \sq	&  	&  	&  	&  	&  	&  	&  	&  	&  	&  	&  	&  	&  	&  	&  	&  	&  	&  	&  	&  	& \sq	&  	&  	&  	&  	&  	&  	&  	&  	&  	&  	& \sq	&  	&  	\\
18.	& BPGC	& \sq	&  	&  	&  	&  	&  	&  	&  	&  	&  	& \cq	&  	& \cq	&  	&  	&  	& \sq	&  	&  	&  	&  	& \cq	&  	&  	& \sq	& \sq	&  	&  	&  	&  	&  	& \sq	& \sq	&  	&  	\\
\rowcolor{white}
19.	& UPB	& \sq	&  	&  	&  	&  	&  	&  	&  	&  	&  	&  	&  	&  	&  	&  	&  	&  	&  	&  	&  	&  	&  	&  	&  	&  	&  	&  	&  	&  	&  	&  	&  	&  	& \sq	&  	\\
20.	& syrapropa	& \sq	&  	&  	&  	&  	&  	&  	&  	&  	&  	&  	&  	&  	&  	&  	&  	& \sq	&  	&  	&  	&  	&  	&  	&  	& \sq	&  	& \sq	&  	&  	&  	&  	&  	&  	&  	& \sq	\\
\rowcolor{white}
21.	& WMD	& \sq	&  	&  	&  	&  	&  	&  	&  	&  	&  	& \cq	&  	&  	& \cq	&  	& \cq	&  	& \cq	&  	&  	&  	& \sq	&  	&  	&  	&  	&  	&  	&  	&  	&  	&  	& \sq	& \sq	&  	\\
22.	& YNUHPCC	& \sq	&  	&  	&  	&  	&  	&  	&  	&  	&  	& \sq	&  	&  	&  	&  	&  	&  	&  	&  	&  	& \cq	&  	&  	&  	&  	&  	&  	&  	&  	&  	&  	&  	&  	&  	&  	\\
\rowcolor{white}
24.	& DoNotDistribute	& \sq	&  	&  	&  	&  	&  	&  	&  	&  	&  	& \sq	&  	&  	&  	&  	&  	&  	&  	&  	&  	&  	&  	&  	& \sq	& \sq	& 	& \sq	&  	&  	&  	&  	&  	&  	&  	&  	\\
25.	& NTUAAILS	&  	&  	&  	&  	&  	&  	&  	&  	&  	&  	& \sq	&  	&  	&  	&  	&  	&  	&  	&  	&  	&  	& \sq	&  	&  	&  	&  	&  	&  	&  	&  	&  	&  	&  	&  	&  	\\
\rowcolor{white}
26.	& UAIC1860	&  	&  	&  	&  	&  	&  	&  	&  	&  	&  	&  	&  	&  	& \cq	& \cq	&  	& \cq	& \sq	&  	&  	&  	& \sq	&  	&  	& \sq	& \sq	&  	&  	&  	&  	&  	&  	&  	&  	&  	\\
27.	& UNTLing	&  	&  	&  	&  	&  	&  	&  	&  	&  	&  	&  	&  	&  	&  	&  	&  	& \sq	&  	&  	&  	&  	& \sq	&  	& \sq	& \sq	&  	&  	&  	&  	& \sq	&  	&  	&  	&  	&  	\\
\bottomrule
\end{tabular}

\begin{tabular}{rl}
1.	& \cite{SemEval20-11-Jurkiewicz}	\\	
2.	& \cite{SemEval20-11-Chernyavskiy}	\\	
3.	& \cite{SemEval20-11-Morio}	\\	
4.	& \cite{SemEval20-11-Raj}		\\	
5.	& \cite{SemEval20-11-Singh}	\\	
6.	& \cite{SemEval20-11-Dimov}		\\	
7.	& \cite{SemEval20-11-Grigorev}	\\	
8.	& \cite{SemEval20-11-Blaschke}	\\	
10.	& \cite{SemEval20-11-Bairaktaris}\\	
\end{tabular}
\hspace{8mm}
\begin{tabular}{rl}
11.	& \cite{SemEval20-11-Kaas}			\\	
13.	& \cite{SemEval20-11-Krishnamurthy}	\\	
14.	& \cite{SemEval20-11-Kim}			\\	
15.	& \cite{SemEval20-11-Altiti}		\\	
16.	& \cite{SemEval20-11-Martinkovic}	\\	
17.	& \cite{SemEval20-11-Jiang}			\\	
18.	& \cite{SemEval20-11-Patil}			\\	
19.	& \cite{SemEval20-11-Paraschiv}		\\	
\\
\end{tabular}
\hspace{8mm}
\begin{tabular}{rl}
20.	& \cite{SemEval20-11-Li}			\\	
21.	& \cite{SemEval20-11-Daval-Frerot}	\\	%
22.	& \cite{SemEval20-11-Dao}			\\	
24.	& \cite{SemEval20-11-Kranzlein}		\\	
25.	& \cite{SemEval20-11-Arsenos}		\\	
26.	& \cite{SemEval20-11-Ermurachi}		\\	
27.	& \cite{SemEval20-11-Petee}			\\	
29.	& \cite{SemEval20-11-Verma}			\\	
\\
\end{tabular}
\caption{Overview of the approaches to the technique classification subtask. 
\sq$=$part of the official submission; \cq$=$considered in internal experiments. 
The references to the description papers appear at the bottom.}
\label{tab:overview_tc}
\end{table}

Team \textbf{Hitachi}\ranktc{3}~\cite{SemEval20-11-Morio} used two distinct FFNs. 
The first one is for sentence representation, whereas the second one is for the 
representation of tokens in the propaganda span. The propaganda span 
representation is obtained by concatenating representation of the 
begin-of-sentence token, span start token, span end token, and aggregated 
representation by attention and max-pooling. As for their winning approach to SI, 
Hitachi trained on the TC subtask independently with different language models
and then combined the resulting models in an ensemble.

As the top-performing models to subtask TC show, whereas the two subtasks can 
be considered as fairly independent, combining them in a reasonable way pays 
back. Additionally, the context of a propaganda snippet is important to 
identify the specific propaganda technique it uses. Indeed, other teams 
tried to make context play a role in their models with certain 
success. For instance, team \textbf{newsSweeper}\ranktc{5}~\cite{SemEval20-11-Singh} 
used RoBERTa to obtain and concatenate representations for both the 
propaganda snippet and its sentence. Team 
\textbf{SocCogCom}\ranktc{13}~\cite{SemEval20-11-Krishnamurthy} reduced the context to 
a window of three words before and after the propaganda snippet.

Again, a number of approaches report sizable improvements when 
adding features. For instance, team \textbf{BPGC}\ranktc{18}~\cite{SemEval20-11-Patil}  
included TF.IDF vectors of words and character $n$-grams, topic modeling, and 
sentence-level polarity (among others) to their ensemble of BERT and logistic 
regression. 
Team \textbf{SocCogCom}\ranktc{13}~\cite{SemEval20-11-Krishnamurthy} integrated 
semantic-level emotional salience features from 
CrystalFeel~\cite{gupta-yang-2018-crystalfeel} and word-level psycholinguistic 
features from LIWC~\cite{LIWC}.
Team \textbf{CyberWallE}\ranktc{8}~\cite{SemEval20-11-Blaschke} 
added named entities, rhetorical, and question features, while taking special 
care of repetitions as part of a complex ensemble architecture.
According to team \textbf{UNTLing}\ranktc{27}~\cite{SemEval20-11-Petee}, considering NEs is 
particularly useful for the \emph{Loaded Language} and the \emph{Flag Waving} 
classes (e.g.,~the latter usually includes references to idealized entities) 
and VAD features were useful for emotion-related techniques such as \emph{Appeal to 
fear/prejudice} and \emph{Doubt}. 
Team \textbf{DiSaster}\ranktc{11}~\cite{SemEval20-11-Kaas} combined BERT with features including frequency of the fragment in the article and in the sentence it appears in, the
inverse uniqueness of words in a span. The goal of the features is to compensate the inability of BERT to deal with distant context, specifically to target the technique \emph{Repetition}. Team \textbf{Solomon}\ranktc{4} also targeted \emph{Repetition} by using dynamic least common sub-sequence is used to score the similarity between the fragment and the context. Then, the fragment is considered to be a repetition if the score is greater than a threshold heuristically set with respect to the length of the fragment. 

Some other teams decided to perform a normalization of the texts, perhaps in 
order to reduce the representation diversity. This was the case of team
\textbf{DUTH}\ranktc{10}~\cite{SemEval20-11-Bairaktaris}, which mapped certain 
words into classes using named entity recognition with focus on person names 
and gazetteers containing names and variations of names of countries (255 
entries), religions (35 entries), political ideologies (23 entries), slogans (41 
entries). The recognized categories were replaced by the category name in the 
input, before passing the input to BERT.

The class distribution for subtask TC is heavily skewed. Whereas team 
\textbf{Inno}\ranktc{7}~\cite{SemEval20-11-Grigorev} experimented with undersampling 
(i.e.~removing some examples from the bigger classes), team 
\textbf{syrapropa}\ranktc{20} applied a cost adjustment to their BERT-based 
model. Team \textbf{UMSIForeseer}\ranktc{17}~\cite{SemEval20-11-Jiang} used a 
mix of oversampling and undersampling, which they combined using a bagging 
ensemble learner.
Finally, some teams decided to apply an overriding strategy on the output of 
their supervised models. Whereas team 
\textbf{CyberWallE}\ranktc{8}~\cite{SemEval20-11-Blaschke} performed a rule-based label 
post-processing, team \textbf{syrapropa}\ranktc{20} applied syntactic rules 
based on part of speech.

\section{Results and Discussion} \label{sec:results}

\subsection{Results on the Span Identification Subtask}

\newcolumntype{G}{>{\columncolor{Gray}}r}
\begin{table}[tbh]
\centering
\begin{tabular}{l|Gggg|Gggg}
\hline 
& \multicolumn{4}{|c}{\bf Test}	& 	\multicolumn{4}{|c}{\bf Development}	\\
\rowcolor{white}
\bf Team  & \bf Rnk & \bf F$_1$		& \bf P	& \bf R		& \bf Rnk&\bf F$_1$	& \bf P		& \bf R	\\ \hline
\rowcolor{white}
Hitachi		& 1& \bf  51.55		& 56.54	& 47.37		& 4	& \bf 50.12	& 42.26		& 61.56	\\
ApplicaAI	& 2	& 49.15		& 59.95	& 41.65		& 3	& 52.19		& 47.15		& 58.44	\\
\rowcolor{white}
aschern		& 3	& 49.10		& 53.23	& 45.56		& 5	& 49.99		& 44.53		& 56.98	\\
LTIatCMU	& 4	& 47.66		& 50.97	& 44.76		& 7	& 49.06		& 43.38		& 56.47	\\
\rowcolor{white}
UPB		& 5	& 46.06		& 58.61	& 37.94		& 8	& 46.79		& 42.44		& 52.13	\\
Fragarach	& 6	& 45.96		& 54.26	& 39.86		& 12	& 44.27		& 41.68		& 47.21	\\
\rowcolor{white}
NoPropaganda	& 7	& 44.68		& 55.62	& 37.34		& 9	& 46.13		& 40.65		& 53.31	\\
CyberWallE	& 8	& 43.86		& 42.16	& 45.70		& 17	& 42.39		& 33.45		& 57.86	\\
\rowcolor{white}
Transformers	& 9	& 43.60		& 49.86	& 38.74		& 14	& 43.06		& 40.85		& 45.52	\\
SWEAT		& 10	& 43.22		& 52.77	& 36.59		& 16	& 42.51		& 42.97		& 42.06	\\
\rowcolor{white}
YNUtaoxin	& 11	& 43.21		& 55.62	& 35.33		& 11	& 44.35		& 40.74		& 48.67	\\
DREAM		& 12	& 43.10		& 54.54	& 35.63		& 19	& 42.15		& 42.66		& 41.65	\\
\rowcolor{white}
newsSweeper	& 13	& 42.21		& 46.52	& 38.63		& 10	& 44.45		& 38.76		& 52.10	\\
PsuedoProp	& 14	& 41.20		& 41.54	& 40.87		& 22	& 39.32		& 34.27		& 46.11	\\
\rowcolor{white}
Solomon		& 15	& 40.68		& 53.95	& 32.66		& 15	& 42.86		& 43.24		& 42.49	\\
YNUHPCC		& 16	& 40.63		& 36.55	& 45.74		& 18	& 42.27		& 32.08		& 61.95	\\
\rowcolor{white}
NLFIIT		& 17	& 40.58		& 50.91	& 33.73		& 21	& 39.67		& 35.04		& 45.72	\\
PALI		& 18	& 40.57		& 53.20	& 32.79		& 2	& 52.35		& 49.64		& 55.37	\\
\rowcolor{white}
UESTCICSA	& 19	& 39.85		& 56.09	& 30.90		& 13	& 44.17		& 43.21		& 45.18	\\
TTUI		& 20	& 39.84	    & \bf 66.88	& 28.37		& 6	& 49.59		& 48.76		& 50.44	\\
\rowcolor{white}
BPGC		& 21	& 38.74		& 49.39	& 31.88		& 25	& 36.79		& 34.72		& 39.12	\\
DoNotDistribute	& 22	& 37.86		& 42.36	& 34.23		& 24	& 37.73		& 32.41		& 45.12	\\
\rowcolor{white}
UTMNandOCAS	& 23	& 37.49		& 37.97	& 37.03		& 31	& 34.35		& 23.65		& 62.69	\\
Entropy		& 24	& 37.23		& 41.68	& 33.63		& 32	& 32.89		& 30.82		& 35.25	\\
\rowcolor{white}
syrapropa	& 25	& 36.20		& 49.53	& 28.52		& 1	& 53.40		& 39.88		& 80.80	\\
SkoltechNLP	& 26	& 34.07		& 46.52	& 26.87		& 26	& 36.70		& 34.99		& 38.59	\\
\rowcolor{white}
NTUAAILS	& 27	& 33.60		& 46.05	& 26.44		& 33	& 31.21		& 27.95		& 35.35	\\
UAIC1860	& 28	& 33.21		& 24.49	& 51.57		& 34	& 30.27		& 20.69		& 56.37	\\
\rowcolor{white}
CCNI		& 29	& 29.48		& 38.09	& 24.05		& 35	& 29.61		& 29.04		& 30.21	\\
NCCU-SMRG	& 30	& 28.47		& 17.30	& \bf 80.37	& 42	& 15.83		& 09.12		& 59.92	\\
\rowcolor{white}
3218IR		& 31	& 23.47		& 22.63	& 24.38		& 40	& 20.03		& 15.10		& 29.76	\\
WMD		& 32	& 20.09		& 47.11	& 12.77		& 27	& 36.34		& 33.26		& 40.05	\\
\rowcolor{white}
LS		& 33	& 18.18		& 34.14	& 12.39		& 29	& 35.49		& 22.41		& \bf 85.33	\\
HunAlize	& 34	& \,\,3.19	& 23.24	& \,\,1.71	& 38	& 24.45		& 37.75		& 18.08	\\
\rowcolor{white}
YOLO		& 35	& \,\,0.72	& 17.20	& \,\,0.37	& 46	& \,\,0.64	& \,\,9.36	& \,\,0.33	\\
Baseline	& 36	& \,\,0.31	& 13.04	& \,\,0.16	& 43	& \,\,1.10	& 11.00		& \,\,0.58	\\
\rowcolor{white}
Murgila		& --	& --		& --	& --		& 20	& 41.38		& 32.96		& 55.56	\\
TakeLab		& --	& --		& --	& --		& 23	& 39.06		& 38.85		& 39.27	\\
\rowcolor{white}
atulcst		& --	& --		& --	& --		& 28	& 36.29		& 38.15		& 34.61	\\
AAA		& --	& --		& --	& --		& 30	& 34.68		& 30.61		& 40.00	\\
\rowcolor{white}
CUNLP		& --	& --		& --	& --		& 36	& 27.78		& \bf 58.23	& 18.24	\\
IIITD		& --	& --		& --	& --		& 37	& 25.82		& 18.81		& 41.15	\\
\rowcolor{white}
UoB		& --	& --		& --	& --		& 39	& 24.02		& 22.30		& 26.04	\\
UBirmingham	& --	& --		& --	& --		& 41	& 16.95		& 23.07		& 13.39	\\
\rowcolor{white}
SocCogCom	& --	& --		& --	& --		& 44	& \,\,0.79	& \,\,9.97	& \,\,0.41	\\
Inno		& --	& --		& --	& --		& 45	& \,\,0.64	& \,\,9.36	& \,\,0.33	\\
\rowcolor{white}
Raghavan	& --	& --		& --	& --		& 47	& \,\,0.40	& \,\,0.20	& 33.45	\\
California	& --	& --		& --	& --		& 48	& \,\,0.39	& \,\,5.92	& \,\,0.20	\\
\hline
\end{tabular}
\caption{\label{tab:span-performance} \textbf{Subtask 1: Span Identification (SI) performance on test and development.} The highest scores for the different measures appear highlighted. 
We found a bug in the evaluation software after the end of the competition. The correct ranking does not differ significantly. The final scores and ranking are available in Appendix~\ref{app:errata}.
} 
\end{table}


Table~\ref{tab:span-performance} shows the performance of the participating 
systems both on the testing and on the development partitions on the SI subtask. 
The baseline for subtask SI is a simple system that randomly generates spans, by first selecting the starting character of a span and then its length. 
As mentioned in Section~\ref{sec:systems}, practically all 
approaches relied on Transformers to produce representations and then plugged 
their output into a sequential model, at the token level. It is worth observing 
that only three of the top-5 systems on the development set appear also 
among the top-5 systems on the test set. Indeed, teams \textbf{syrapropa} 
and \textbf{PALI} felt down from positions 1 and 2 on development to positions 25 and 18 on test, which suggests possible overfitting. The 
performance for the final top-3 systems on the test partition 
---\textbf{Hitachi}, \textbf{ApplicaAI}, and \textbf{aschern}--- reflects 
robust systems that generalize much better. 

\begin{figure}
  \centering
  \includegraphics[width=1.02\textwidth]{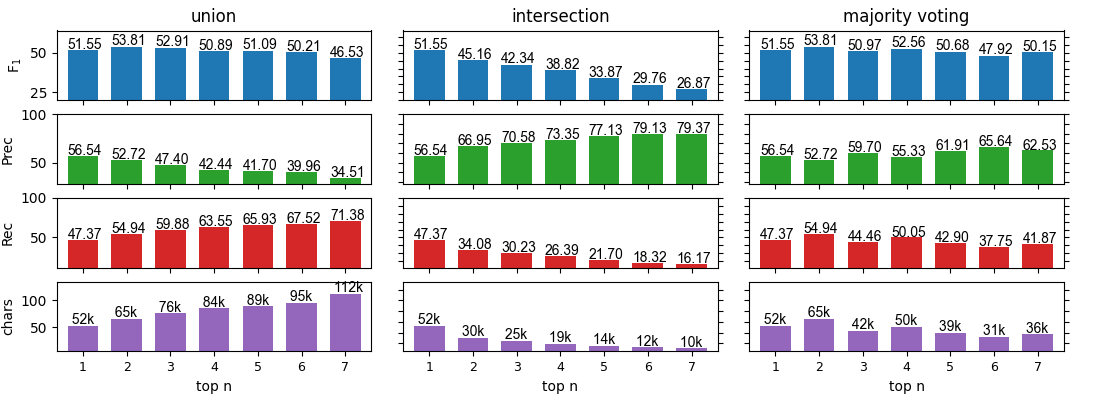}
  \caption{\textbf{SI Subtask}: performance when combining the top-7 systems 
  using union, intersection, and majority voting. The bottom plots 
  show the number of characters deemed propaganda in each combination.}
  \label{fig:eval_si_combinations}
\end{figure}

Figure~\ref{fig:eval_si_combinations} shows the performance evolution when 
combining the top-performing systems on the test set.  All 
operations are carried out at the character level. Union and intersection 
involve the corresponding set operations. In union, a character is considered 
as propaganda if at least one of the systems involved has recognized it as part 
of a propaganda snippet. In intersection, a character is considered as 
propaganda if all systems have flagged it. For majority voting, we consider a 
character as propaganda if more than 50\% of the systems involved had flagged 
it. The precision and the recall trends are just as expected: a lower precision 
(higher recall) is observed when more systems are combined with a union 
operation, and the opposite occurs for the intersection. Despite the loss in 
terms of precision, computing the union of the top-$[2,3]$ systems results in a 
better performance than the top system in isolation. Such a combination gathers 
large ensembles of Transformer representations together with self-supervision to 
produce additional training data and boundary post-processing. If we are 
interested in a high-precision model, applying the intersection would make more sense, as it reaches a precision of 66.95 when combining the top-2 systems. 
Nevertheless, the amount of snippets lost is significant, causing the recall to 
drop accordingly. 
The majority voting lies in between: keeping reasonable values of both 
precision and recall.

\begin{figure}[t]
  \hspace{-8mm}
  \includegraphics[width=1.07\textwidth]{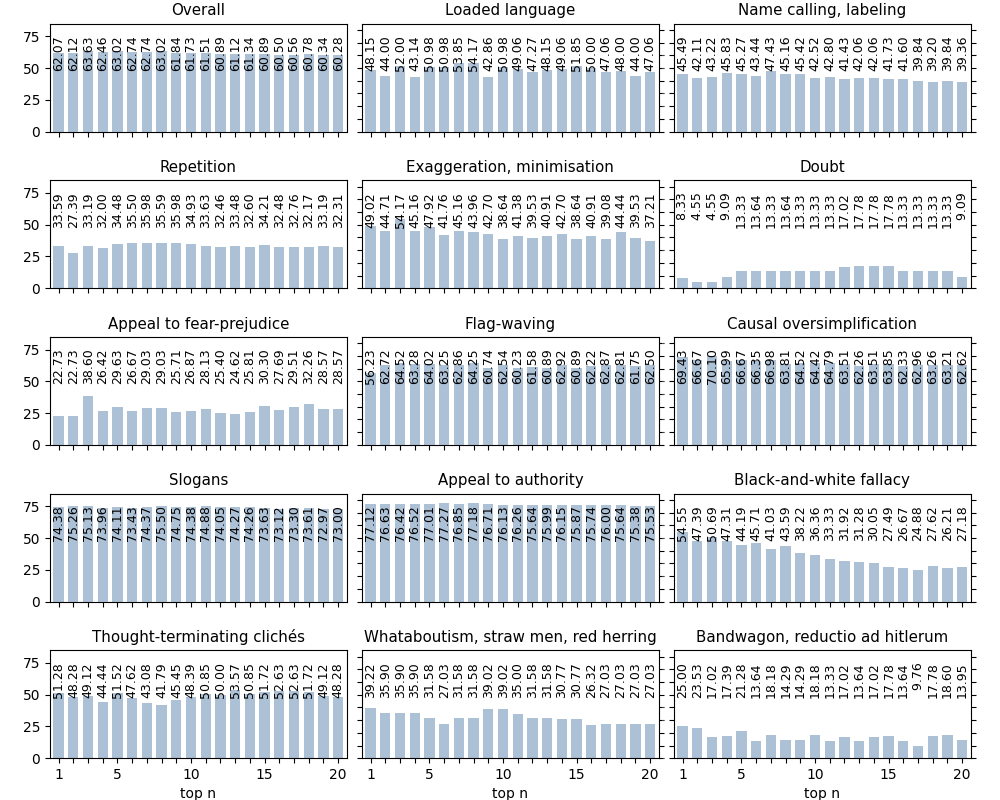}
  \caption{F$_1$ performance for the technique classification subtask when combining 
  up to top-20 systems with majority voting. The plots show the overall 
  performance (top left) as well as for each of the 14 classes.}
  \label{fig:eval_tc_combinations}
\end{figure}

\subsection{Results on the Technique Classification Subtask}

Table~\ref{tab:technique-test} shows the performance of the participating 
systems on the test set for the TC subtask, while Table~\ref{tab:technique-dev} reports the results on the development set. 
The baseline system for subtask TC is a logistic regression classifier using one feature only: the length of the fragment.  
A similar pattern as for the SI subtask is observed: only 
two of the top-5 systems on the development set appear also 
among the top-5 systems on the test set. At the same time, systems that appeared to have a modest performance on the development set could eventually reach a higher position on test. For instance, team 
\textbf{Hitachi}, which was ranked 8th on development, ended up in the third position on the test set. 

Beside the performance on the full dataset, the tables show the performance for 
each of the target 14 classes. In general, the systems show reasonably good performance when predicting \emph{Loaded 
Language} and \emph{Name Calling or Labeling}. These two classes are the most frequent 
ones, by a margin, and are also among the shortest ones on average 
(cf.~Figure~\ref{fig:statscorpus}). On the other hand, techniques 13 (\emph{Straw man, red herring}) and 14 (\emph{Bandwagon, reduction ad hitlerum, whataboutism}) are among the 
hardest to identify. They are also among the least frequent ones. 

Once again, we studied the performance when combining more approaches. 
Figure~\ref{fig:eval_tc_combinations} shows the performance evolution when 
combining different numbers of top-performing systems on the test set. 
As this is a multi-class problem, we combine the systems only on the basis of 
majority voting. In case of a tie, we prefer the more 
frequent technique on the training set. When looking at the overall picture, the 
performance evolution when adding more systems is fairly flat, reaching the 
top performance when combining the top-3 systems: 63.63, which represents more 
than 1.5 points of improvement absolute over the top-1 system. When zooming 
into each of the fourteen classes, we observe that in general the performance peak is 
indeed reached when considering three systems, e.g.,~for \emph{Appeal to 
fear--prejudice}, \emph{Exaggeration, minimisation}, or \emph{Causal oversimplification}. 
Still for \emph{Doubt}, which is the hardest class to recognize, as many as 13 systems are 
necessary to reach a (still discrete) peak performance of 17.78.  There 
are other classes, such as \emph{Black-and-white fallacy} or \emph{Whatabaotism, straw 
men, red herring}, for which system combinations do not help. 

\begin{landscape}
\begin{table}
\renewcommand{\arraystretch}{0.94}%
\begin{tabular}{rl| G |GGGGGGG GGGGGGG}
\rowcolor{white}
Rnk & Team	& Overall	& 1	& 2	& 3	& 4	& 5	& 6	& 7	& 8	& 9	& 10	& 11	& 12	& 13	& 14	\\
\hline 
\rowcolor{white}
1  & ApplicaAI&\bf 62.07& 77.12	& 74.38	&\bf 54.55&33.59& 56.23&\bf 45.49&\bf 69.43& 22.73&51.28& 48.15	& 49.02	& 39.22	& 25.00	& 8.33	\\
2  & aschern	& 62.01	& 77.02	&\bf 75.65&53.38& 32.65	& 59.44	& 41.78	& 66.35	& 25.97	&\bf 54.24&35.29&\bf 53.57&\bf 42.55& 18.87& 14.93	\\
\rowcolor{white}
3  & Hitachi	& 61.73	& 75.64	& 74.20	& 37.88	& 34.58	&\bf 63.43& 38.94&68.02	&\bf 36.62&45.61& 40.00	& 47.92	& 29.41	&\bf 26.92 & 4.88	\\
4  & Solomon	& 58.94	& 74.66	& 70.75	& 42.53	& 28.44	& 61.82	& 39.39	& 61.84	& 19.61	& 50.75	& 26.67	& 42.00	& 38.10	& 0.00	& 4.88	\\
\rowcolor{white}
5  &newsSweeper	& 58.44	& 75.32	& 74.23	& 20.69	& 37.10	& 56.55	& 42.80	& 60.53	& 19.72	& 50.75	& 41.67	& 25.00	& 21.62	& 8.00	& 13.04	\\
6  &NoPropaganda& 58.27	&\bf 77.17&73.90& 42.71	&\bf 37.99&56.27& 38.02	& 59.30	& 12.12	& 42.42	& 23.26	& 8.70	& 23.26	& 0.00	& 0.00	\\
\rowcolor{white}
7  & Inno	& 57.99	& 73.31	& 74.30	& 24.89	& 35.39	& 58.65	& 45.09	& 59.41	& 24.32	& 43.75	& 43.14	& 40.40	& 29.63	& 19.36	& 10.71	\\
8  & CyberWallE	& 57.37	& 74.68	& 70.92	& 47.68	& 28.34	& 58.65	& 39.84	& 54.38	& 15.39	& 39.39	& 14.63	& 23.68	& 23.81	& 0.00	& 12.25	\\
\rowcolor{white}
9  & PALI	& 57.32	& 74.29	& 69.09	& 24.56	& 28.57	& 58.97	& 36.59	& 61.62	& 30.59	& 39.22	& 27.59	& 39.62	& 40.82	& 20.90	&\bf 28.57	\\
10 & DUTH	& 57.21	& 73.71	& 71.41	& 20.10	& 28.24	& 59.16	& 33.33	& 58.95	& 26.23	& 34.78	& 44.44	& 33.33	& 27.03	& 17.78	& 9.30	\\
\rowcolor{white}
11 & DiSaster	& 56.65	& 74.49	& 68.10	& 20.44	& 30.64	& 59.12	& 35.25	& 58.25	& 14.63	& 42.55	& 51.16	& 26.67	& 19.05	& 4.35	& 20.41	\\
12 & djichen	& 56.54	& 73.21	& 68.38	& 29.75	& 31.42	& 60.00	& 33.65	& 56.19	& 22.79	& 30.77	& 37.50	& 43.81	& 27.91	& 18.87	& 20.83	\\
\rowcolor{white}
13 & SocCogCom	& 55.81	& 72.18	& 67.34	& 18.88	& 34.86	& 60.40	& 31.62	& 54.26	& 6.35	& 40.91	& 28.57	& 26.51	& 23.53	& 10.00	& 9.76	\\
14 & TTUI	& 55.64	& 73.22	& 68.49	& 21.18	& 32.20	& 57.40	& 41.48	& 61.68	& 23.08	& 37.50	& 28.24	& 35.29	& 25.00	& 20.29	& 24.56	\\
\rowcolor{white}
15 & JUST	& 55.31	& 71.96	& 64.73	& 21.94	& 29.57	& 58.26	& 37.10	& 62.56	& 27.27	& 33.33	&\bf 48.89&28.89& 31.82	& 28.57	& 24.49	\\
16 & NLFIIT	& 55.25	& 72.55	& 69.30	& 21.55	& 30.30	& 55.66	& 24.89	& 63.32	& 0.00	& 41.67	& 29.63	& 32.10	& 13.64	& 0.00	& 9.30	\\
\rowcolor{white}
17 &UMSIForeseer& 55.14	& 73.02	& 70.79	& 21.49	& 28.57	& 57.21	& 31.97	& 56.14	& 0.00	& 39.22	& 29.41	& 0.00	& 14.29	& 0.00	& 9.76	\\
18 & BPGC	& 54.81	& 71.58	& 67.51	& 23.74	& 33.47	& 53.78	& 33.65	& 58.93	& 24.18	& 40.00	& 30.77	& 40.00	& 20.69	& 20.90	& 12.50	\\
\rowcolor{white}
19 & UPB	& 54.30	& 70.09	& 68.86	& 20.00	& 30.62	& 52.55	& 30.00	& 55.87	& 16.95	& 34.62	& 20.00	& 19.72	& 22.86	& 4.88	& 0.00	\\
20 & syrapropa	& 54.25	& 71.47	& 68.44	& 30.77	& 28.10	& 56.14	& 29.77	& 57.02	& 21.51	& 29.03	& 31.58	& 30.61	& 28.57	& 9.09	& 19.61	\\
\rowcolor{white}
21 & WMD	& 52.01	& 69.33	& 64.67	& 13.89	& 25.46	& 53.94	& 29.20	& 52.08	& 5.71	& 6.90	& 7.14	& 0.00	& 7.41	& 0.00	& 5.00	\\
22 & YNUHPCC	& 50.50	& 68.08	& 62.33	& 17.72	& 21.54	& 51.04	& 26.40	& 55.56	& 3.45	& 27.59	& 29.79	& 38.38	& 17.78	& 15.00	& 13.79	\\
\rowcolor{white}
23 & UESTCICSA	& 49.94	& 68.23	& 66.88	& 27.96	& 25.44	& 44.99	& 22.75	& 53.14	& 3.74	& 41.38	& 12.77	& 11.27	& 28.57	& 3.70	& 0.00	\\
24&DoNotDistribute&49.72& 68.44	& 60.65	& 19.44	& 27.23	& 46.25	& 29.75	& 53.76	& 14.89	& 28.07	& 22.64	& 24.49	& 12.25	& 9.68	& 4.55	\\
\rowcolor{white}
25 & NTUAAILS	& 46.37	& 65.79	& 54.55	& 18.43	& 29.66	& 48.75	& 28.31	& 46.47	& 0.00	& 13.79	& 36.36	& 0.00	& 11.43	& 4.08	& 9.76	\\
26 & UAIC1860	& 41.17	& 62.33	& 42.97	& 11.16	& 21.01	& 36.41	& 22.12	& 38.78	& 7.60	& 11.43	& 17.39	& 2.90	& 5.56	& 4.26	& 9.76	\\
\rowcolor{white}
27 & UNTLing	& 39.11	& 62.57	& 36.74	& 7.78	& 11.82	& 32.65	& 5.29	& 40.48	& 2.86	& 17.65	& 4.35	& 0.00	& 0.00	& 0.00	& 0.00	\\
28 & HunAlize	& 37.10	& 58.59	& 15.82	& 2.09	& 23.81	& 31.76	& 11.83	& 29.95	& 7.84	& 4.55	& 6.45	& 8.00	& 0.00	& 0.00	& 0.00	\\
\rowcolor{white}
29 &Transformers& 26.54	& 47.55	& 24.06	& 2.86	& 0.00	& 0.98	& 0.00	& 0.00	& 0.00	& 0.00	& 0.00	& 0.00	& 0.00	& 0.00	& 0.00	\\
30 & Baseline	& 25.20	& 46.48	& 0.00	& 19.26	& 14.42	& 29.14	& 3.68	& 6.20	& 11.56	& 0.00	& 0.00	& 0.00	& 0.00	& 0.00	& 0.00	\\
\rowcolor{white}
31 & Entropy	& 20.39	& 37.74	& 15.49	& 5.83	& 6.39	& 12.81	& 6.32	& 4.95	& 7.41	& 0.00	& 3.92	& 2.27	& 0.00	& 6.78	& 0.00	\\
32 & IJSE8	& 19.72	& 38.07	& 14.70	& 4.92	& 8.23	& 15.47	& 7.07	& 8.57	& 2.27	& 0.00	& 0.00	& 0.00	& 0.00	& 0.00	& 0.00	\\
\end{tabular}

\caption{\label{tab:technique-test} \textbf{Technique classification F$_1$ performance on the test set}. The systems are ordered on the basis of the final 
ranking. Columns \textbf{1} to \textbf{14} show the performance on each of the 
fourteen classes (cf.\ Section~\ref{sec:propagandatechniques}). The best score 
on each technique appears highlighted. We found a bug in the evaluation software after the end of the competition. The correct ranking does not differ significantly. The final scores and ranking are available in Appendix~\ref{app:errata}. 
}

\end{table}
\end{landscape}

\begin{landscape}
\begin{table}
\renewcommand{\arraystretch}{0.87}%
\footnotesize
\begin{tabular}{rl| G |GGGGGGG GGGGGGG}
\rowcolor{white}
 Rnk	& Team	& Overall	& 1	& 2	& 3	& 4	& 5	& 6	& 7	& 8	& 9	& 10	& 11	& 12	& 13	& 14
	\\
\hline 
\rowcolor{white}
1	& ApplicaAI &\bf 70.46	& 80.42	& 74.11	& 70.67	&\bf 60.93& 63.57&\bf 48.49&\bf 82.56&47.06& 62.65&\bf 42.86&20.00& 43.48&\bf 41.67& 80.00	\\
2	& aschern	& 68.11	& 80.00	& 73.85	& 67.39	& 59.36	&\bf 66.19&45.24& 74.25	& 30.44	& 63.89	& 28.57	& 32.43	& 33.33	& 38.46	& 40.00	\\
\rowcolor{white}
8	& Hitachi	& 65.19	& 77.99	& 71.16	& 46.98	& 56.58	& 62.96	& 36.96	& 81.40	& 50.00	& 67.61	& 12.50	& 22.22	& 19.05	& 40.91	& 75.00	\\
16	& Solomon	& 59.55	& 75.92	& 71.98	& 33.48	& 49.65	& 51.28	& 39.08	& 74.68	& 34.29	& 45.90	& 0.00	& 20.83	& 32.26	& 20.83	& 0.00  \\
\rowcolor{white}
5	& FakeSolomon	& 67.17	& 79.29	& 74.38	& 66.04	& 50.75	& 61.22	& 42.72	& 74.36	& 31.25	&\bf 70.00&9.52	& 18.75	& 43.48	& 6.25	&\bf 88.89	\\
10	& newsSweeper	& 62.75	& 76.23	& 71.35	& 45.14	& 55.32	& 56.94	& 44.90	& 74.29	& 47.83	& 60.53	& 16.67	& 6.90	& 16.67	& 6.45	& 57.14	\\
\rowcolor{white}
11	& NoPropaganda	& 60.68	& 77.06	& 72.49	& 47.69	& 38.71	& 50.36	& 33.33	& 75.74	& 48.00	& 53.17	& 0.00	& 0.00	& 8.33	& 0.00	& 0.00	\\
13	& Inno		& 60.11	& 76.57	& 70.62	& 34.67	& 43.75	& 52.86	& 40.82	& 77.97	& 34.15	& 59.70	& 11.77	& 35.71	& 18.18	& 15.39	& 61.54	\\
\rowcolor{white}
7	& CyberWallE	& 66.42	& 76.62	&\bf 81.00&\bf 73.29&52.70& 53.85& 30.61& 73.68	& 21.05	& 51.35	& 18.18	& 21.43	& 17.39	& 0.00	& 22.22	\\
6	& PALI		& 66.89	& 78.01	& 76.34	& 61.59	& 48.61	& 58.46	& 40.45	& 79.53	& 42.11	& 58.33	& 41.67	& 35.56	&\bf 45.71&27.03& 75.00	\\
\rowcolor{white}
23	& DUTH		& 57.86	& 76.12	& 70.32	& 23.08	& 46.15	& 48.72	& 34.41	& 64.43	& 42.86	& 55.39	& 9.52	& 11.11	& 15.39	& 11.11	& 50.00	\\
9	& DiSaster	& 62.84	& 76.73	& 77.33	& 47.11	& 48.28	& 56.21	& 35.29	& 75.15	& 23.08	& 35.09	& 30.00	& 0.00	& 9.09	& 5.56	& 28.57	\\
\rowcolor{white}
26	& djichen	& 57.57	& 76.04	& 70.53	& 24.89	& 44.60	& 48.28	& 36.78	& 76.36	& 31.82	& 51.43	& 13.33	& 20.00	& 10.81	& 4.55	& 0.00	\\
28	& SocCogCom	& 57.01	& 70.65	& 64.38	& 31.78	& 45.67	& 53.99	& 32.91	& 77.11	& 28.57	& 30.19	& 0.00	& 21.43	& 12.90	& 6.45	& 57.14	\\
\rowcolor{white}
17	& TTUI		& 58.98	& 75.08	& 71.00	& 32.07	& 41.96	& 56.92	& 34.69	& 77.53	& 33.33	& 50.00	& 16.22	& 19.05	& 20.69	& 29.09	& 66.67	\\
24	& JUST		& 57.67	& 74.97	& 72.36	& 23.64	& 49.64	& 48.75	& 31.07	& 73.75	& 23.81	& 29.03	& 0.00	& 0.00	& 0.00	& 5.88	& 0.00	\\
\rowcolor{white}
20	& NLFIIT	& 58.51	& 74.46	& 70.48	& 33.33	& 41.48	& 53.13	& 35.14	& 75.00	& 34.29	& 60.53	& 13.33	& 27.78	& 24.24	& 24.14	& 60.00	\\
19	& UMSIForeseer	& 58.70	& 74.44	& 70.95	& 31.76	& 51.80	& 47.95	& 31.58	& 75.86	&\bf 52.63&43.33& 10.00	& 0.00	& 0.00	& 10.00	& 0.00	\\
\rowcolor{white}
18	& BPGC		& 58.80	& 75.45	& 70.05	& 29.75	& 49.64	& 52.86	& 32.43	& 75.82	& 27.03	& 52.31	& 19.05	& 18.75	& 23.08	& 9.52	& 0.00	\\
21	& UPB		& 58.33	& 74.02	& 71.64	& 24.16	& 50.00	& 46.15	& 29.89	& 70.73	& 34.29	& 50.00	& 7.41	& 8.70	& 0.00	& 6.06	& 57.14	\\
\rowcolor{white}
3	& syrapropa	& 67.83	& 77.24	& 77.00	& 70.12	& 51.39	& 56.74	& 40.82	& 78.89	& 41.86	& 68.49	& 20.00	& 13.79	& 26.09	& 24.24	& 60.00	\\
33	& WMD		& 52.31	& 71.57	& 57.70	& 40.00	& 35.12	& 39.77	& 26.83	& 63.10	& 6.90	& 15.69	& 19.05	& 0.00	& 0.00	& 0.00	& 40.00	\\
\rowcolor{white}
29	& YNUHPCC	& 56.16	& 70.57	& 69.45	& 29.75	& 36.36	& 51.39	& 26.09	& 73.05	& 27.27	& 40.58	& 6.45	& 18.18	& 26.09	& 21.05	& 54.55	\\
25	& UESTCICSA	& 57.57	& 74.15	& 65.04	& 48.37	& 39.66	& 46.48	& 38.71	& 63.23	& 26.67	& 58.82	& 0.00	& 0.00	& 36.36	& 9.52	& 0.00	\\
\rowcolor{white}
30   & DoNotDistribute	& 54.00	& 71.20	& 67.33	& 29.38	& 31.93	& 45.95	& 25.58	& 72.61	& 25.00	& 28.99	& 0.00	& 0.00	& 22.86	& 9.76	& 0.00	\\
32	& NTUAAILS	& 53.25	& 69.76	& 59.90	& 27.19	& 43.64	& 51.61	& 21.33	& 73.03	& 17.14	& 27.12	& 6.67	& 7.14	& 0.00	& 8.00	& 44.44	\\
\rowcolor{white}
34	& UAIC1860	& 43.84	& 57.88	& 40.37	& 14.12	& 23.66	& 42.03	& 7.02	& 60.00	& 0.00	& 4.55	& 10.53	& 0.00	& 0.00	& 0.00	& 33.33	\\
37	& UNTLing	& 40.92	& 59.45	& 33.33	& 11.83	& 12.28	& 35.42	& 11.94	& 51.70	& 26.67	& 25.00	& 8.33	& 0.00	& 0.00	& 5.88	& 0.00	\\
\rowcolor{white}
36	& HunAlize	& 41.02	& 56.13	& 40.75	& 5.32	& 22.22	& 41.18	& 9.23	& 53.50	& 33.33	& 0.00	& 0.00	& 0.00	& 0.00	& 0.00	& 0.00	\\
41	& Transformers	& 30.10	& 48.50	& 28.62	& 12.31	& 0.00	& 0.00	& 0.00	& 0.00	& 0.00	& 0.00	& 0.00	& 0.00	& 0.00	& 4.88	& 0.00	\\
\rowcolor{white}
43	& Baseline	& 26.53	& 40.58	& 0.00	& 38.50	& 11.68	& 19.20	& 9.38	& 8.28	& 7.23	& 0.00	& 0.00	& 0.00	& 0.00	& 0.00	& 0.00	\\
27	& Entropy	& 57.10	& 75.22	& 66.50	& 28.19	& 45.16	& 53.60	& 31.11	& 74.85	& 29.27	& 46.88	& 13.33	& 18.18	& 20.00	& 10.00	& 0.00	\\
\rowcolor{white}
4	& FLZ		& 67.17	&\bf 81.09&71.12& 70.78	& 23.38	& 65.46	& 20.00	& 64.08	& 46.15	& 64.00	& 44.44	&\bf 37.04&28.57& 6.67	& 75.00	\\
12	& Fragarach	& 60.49	& 77.12	& 70.50	& 38.21	& 50.69	& 51.03	& 40.91	& 74.12	& 34.04	& 45.71	& 14.29	& 12.90	& 16.67	& 6.06	& 57.14	\\
\rowcolor{white}
14	& NerdyBirdies	& 59.83	& 76.23	& 73.55	& 33.05	& 48.65	& 52.06	& 36.04	& 70.30	& 47.62	& 48.57	& 15.39	& 8.70	& 20.00	& 13.33	& 0.00	\\
15	& CUNLP		& 59.55	& 74.38	& 73.30	& 47.62	& 43.66	& 50.69	& 34.29	& 71.90	& 35.90	& 55.07	& 16.22	& 15.79	& 19.51	& 8.89	& 25.00	\\
\rowcolor{white}
22	& Tianyi	& 58.23	& 75.39	& 71.22	& 26.43	& 44.72	& 50.36	& 34.57	& 75.61	& 28.57	& 49.32	& 13.33	& 11.77	& 0.00	& 0.00	& 57.14	\\
31	& Murgila	& 53.53	& 71.21	& 66.14	& 24.55	& 42.55	& 37.29	& 22.47	& 69.51	& 30.00	& 40.58	& 17.65	& 13.33	& 31.58	& 22.64	& 44.44	\\
\rowcolor{white}
35	& hseteam	& 41.40	& 63.65	& 42.82	& 22.40	& 32.26	& 26.29	& 11.11	& 57.52	& 7.69	& 26.67	& 0.00	& 7.27	& 0.00	& 4.76	& 42.86	\\
38 & HenryAtDuderstadt	& 34.15	& 52.97	& 4.10	& 0.00	& 0.00	& 34.29	& 0.00	& 0.00	& 0.00	& 0.00	& 0.00	& 0.00	& 0.00	& 0.00	& 0.00	\\
\rowcolor{white}
39 & XJPaccelerationMASTER& 32.46& 48.76& 0.00	& 0.00	& 0.00	& 33.06	& 0.00	& 0.00	& 0.00	& 0.00	& 0.00	& 0.00	& 0.00	& 0.00	& 0.00	\\
40	& SWEAT		& 30.57	& 46.83	& 0.00	& 0.00	& 0.00	& 0.00	& 0.00	& 0.00	& 0.00	& 0.00	& 0.00	& 0.00	& 0.00	& 0.00	& 0.00	\\
\rowcolor{white}
42	& NCCU-SMRG	& 29.26	& 37.92	& 34.46	& 25.53	& 14.69	& 36.92	& 15.09	& 45.07	& 9.76	& 13.33	& 12.50	& 16.67	& 12.50	& 0.00	& 23.08	\\
44	& LS		& 26.23	& 39.20	& 0.00	& 39.31	& 12.95	& 19.67	& 9.09	& 8.28	& 8.79	& 0.00	& 0.00	& 0.00	& 6.25	& 0.00	& 0.00	\\
\rowcolor{white}
45	& SkoltechNLP	& 26.23	& 39.20	& 0.00	& 39.31	& 12.95	& 19.67	& 9.09	& 8.28	& 8.79	& 0.00	& 0.00	& 0.00	& 6.25	& 0.00	& 0.00	\\
46	& TakeLab	& 25.31	& 42.92	& 24.70	& 1.37	& 0.00	& 6.43	& 0.00	& 0.00	& 0.00	& 0.00	& 0.00	& 0.00	& 0.00	& 0.00	& 0.00	\\
\rowcolor{white}
47	& UTMNandOCAS	& 23.42	& 40.53	& 19.74	& 18.67	& 4.44	& 0.00	& 10.53	& 0.00	& 0.00	& 0.00	& 0.00	& 23.08	& 8.70	& 0.00	& 0.00	\\
\end{tabular}
\caption{\label{tab:technique-dev} \textbf{Technique classification F$_1$ performance on the development set}. The systems are ordered based on the final ranking on the test set (cf.\ Table~\ref{tab:technique-test}), whereas the ranking is the one on the development set.  Columns \textbf{1} to \textbf{14} show the performance on each class (cf.\ Section~\ref{sec:propagandatechniques}). The best score on each class is marked.}

\end{table}
\end{landscape}

\section{Related Work} 
\label{sec:pilot}

Propaganda is particularly visible in the context of ``fake news'' on social media, which have attracted a lot of research recently \cite{Shu:2017:FND:3137597.3137600}. \newcite{thorne-vlachos:2018:C18-1} surveyed fact-checking approaches to fake news and related problems, and \newcite{Li:2016:STD:2897350.2897352} focused on truth discovery in general. Two recent articles in \emph{Science} offered a general discussion on the science of ``fake news'' \cite{Lazer1094} and the process of proliferation of true and false news online \cite{Vosoughi1146}. 

We are particularly interested here in how different forms of propaganda are manifested in text.
So far, the computational identification of propaganda has been tackled mostly at the article level. \newcite{rashkin-EtAl:2017:EMNLP2017} created a corpus, where news articles are labeled as belonging to one of four categories: {\em propaganda}, {\em trusted}, {\em hoax}, or {\em satire}. The articles came from eight sources, two of which are considered propaganda. The labels were obtained using distant supervision, assuming that all articles from a given news source share the label of that source, which inevitably introduces noise~\cite{Horne2018}. \newcite{BarronIPM:19} experimented with a binary version of the problem: {\em propaganda} vs.\ {\em no propaganda}.

In general, propaganda techniques serve as a means to persuade people, often in argumentative settings. While they may increase the rhetorical effectiveness of arguments, they naturally harm other aspects of argumentation quality \cite{wachsmuth:2017}. In particular, many of the span propaganda techniques considered in this shared task relate to the notion of {\em fallacies}, i.e.~arguments whose reasoning is flawed in some way, often hidden and often on purpose \cite{tindale:2007}. Some recent work in computational argumentation have dealt with such fallacies. Among these, \newcite{habernal:2018a} presented and analyzed a corpus of web forum discussions with \textit{Ad hominem} fallacies, and \newcite{Habernal.et.al.2017.EMNLP} introduced \textit{Argotario}, a game that educates people to recognize fallacies. Argotario also had a corpus as a by-product, with 1.3k arguments annotated for five fallacies, including \textit{Ad hominem}, \textit{Red herring} and \textit{Irrelevant authority}, which directly relate to propaganda techniques (cf.\ Section~\ref{sec:propagandatechniques}). Different from these corpora, the news articles in our corpus have 14~different techniques annotated. Instead of labeling entire arguments, our annotations aim at identifying the minimal text spans related to a technique.   

We already used the news articles included in our corpus previously in a pilot task that ran in January 2019, the {\em Hack the News Datathon},\footnote{\url{https://www.datasciencesociety.net/hack-news-datathon/}}
as well as in a previous shared task, held as part of the {\em 2019 Workshop on NLP4IF: Censorship, Disinformation, and Propaganda}.\footnote{\url{http://www.netcopia.net/nlp4if/2019/}} 
Both the datathon and the shared task tackled the identification of propaganda techniques as one overall task (along with a binary sentence-level propaganda classification task), i.e.~without splitting it into subtasks. As detailed in the overview paper of~\newcite{da-san-martino-etal-2019-findings}, the best-performing models in the shared task used BERT-based contextual representations. Others used contextual representations based on RoBERTa, Grover, and ELMo, or context-independent representations based on lexical, sentiment, readability, and TF-IDF features. As in the task at hand, ensembles were also popular.  Still, the most successful submissions achieved an F$_1$-score of 24.88 only (and only 10.43 in the datathon). As a result, we decided to split the task into subtasks in order to allow researchers to focus on one subtask at a time. Moreover, we merged some of the original 18 propaganda techniques to help with data sparseness issues.

Other related shared tasks include the FEVER 2018 and 2019 tasks on {\em Fact Extraction and VERification}~\cite{thorne-EtAl:2018:N18-1}, and the SemEval 2019 task on {\em Fact-Checking in Community Question Answering Forums}~\cite{mihaylova-etal-2019-semeval}. Also, the CLEF {\em CheckThat!} labs' shared tasks \cite{clef2018checkthat:overall,clef-checkthat:2019,CheckThat:ECIR2019} featured challenges on automatic identification \cite{clef2018checkthat:task1,clef-checkthat-T1:2019} and verification \cite{clef2018checkthat:task2,clef-checkthat-T2:2019} of claims in political debates in the past few years.

\section{Conclusions and Future Work}
\label{sec:conclusions}

We have described the SemEval-2020 Task 11 on Detection of Propaganda Techniques in News Articles. 
The task attracted the interest of a number of researchers: 250 teams signed up to participate, and 44 made submissions on test. 
We received 35 and 31 submissions for subtask SI and subtask TC, respectively. Overall, subtask SI (segment identification) was easier and all systems managed to improve over the baseline. However, subtask TC (technique classification) proved to be much more challenging, and some teams could not improve over our baseline.

In future work, we plan to extend the dataset to cover more examples 
as well as more propaganda techniques. We further plan to develop similar datasets for other languages.

As a final note, we would like to warn about the ethical consequences of deploying automatic propaganda detection in practice, e.g., a respective system may detect propaganda falsely. We see automatic propaganda detection as a tool to raise awareness and to educate users in spotting the use of propaganda in the news by themselves, i.e.,~without the need to rely on such systems. To this end, a system should also make it clear that not everything that appears to be propaganda actually is.

\pn{\section*{Acknowledgments}
This research is part of the Tanbih project,\footnote{\url{http://tanbih.qcri.org/}} which aims to limit the effect of ``fake news'', propaganda and media bias by making users aware of what they are reading.}

\bibliographystyle{coling}
\bibliography{semeval20-task11,propaganda,other,semeval2020}

\appendix

\section{Summary of all Submitted Systems}
\label{app:full_description}

This appendix includes brief summaries of all the teams' approaches to both subtasks. We sort the teams in alphabetical order. The subindex on the right of each team represents its official test rank in the subtasks. Teams appearing in Tables~\ref{tab:span-performance}, \ref{tab:technique-test}, or~\ref{tab:technique-dev} but not here did not submit a paper describing their approach. 

\paragraph{Team 3218IR~\cite{SemEval20-11-Dewantara}\ranksi{31}}  employed a one-dimensional CNN with word embeddings, whose number of layers and filters as well as kernel and pooling sizes are all tuned empirically. 

\paragraph{Team ApplicaAI \cite{SemEval20-11-Jurkiewicz}\rankboth{2}{1}}  applied self-supervision using the RoBERTa model. For the SI subtask, they used a RoBERTa-CRF architecture. The model trained using this architecture was then iteratively used to produce silver data by predicting on 500k sentences and retraining the model with both gold and silver data. As for subtask TC, ApplicaAI opted for feeding their models with propagandas snippets in context. Full sentences are shaped as the input with the specific propaganda in them. Once again, silver data was used, taking advantage of the snippets spotted by their SI model and labeling with their preliminary TC model. The final classifier was an ensemble of models trained on the original corpus, re-weighting, and a model trained also on silver data. 

\paragraph{Team aschern~\cite{SemEval20-11-Chernyavskiy}\rankboth{3}{2}} tackled both subtasks. For SI, the system fine-tunes an ensemble of two differently intialized RoBERTa models, each with an attached CRF for sequence labeling and simple span character boundary post-processing. A RoBERTa ensemble is also used for TC, treating the task as sequence classification but using an average embedding of the surrounding tokens and the length of a span as contextual features. Via transfer learning, knowledge from both tasks is incorporated. Finally, specific postprocessing is done to increase the consistency of the repetition technique spans and to avoid insertions of techniques in other techniques. 

\paragraph{Team BPGC~\cite{SemEval20-11-Patil}\rankboth{21}{18}} used a multigranularity approach to address subtask SI. Information at the full article and sentence level was considered when classifying each word as propaganda or not, by means of computing and concatenating vectorial representations for the three inputs. For subtask TC they used an ensemble of BERT and logistic regression classifiers, complemented with engineered features which, as stated by the authors, were particularly useful for minority classes. Such features include TF-IDF vectors of words and character $n$-grams, topic modeling, and sentence-level polarity, among others. Different learning models were explored for both tasks, including LSTM and CNN, together with diverse transformers to build ensembles of classifiers.

\paragraph{Team CyberWallE~\cite{SemEval20-11-Blaschke}\rankboth{8}{8}} used BERT embeddings for subtask SI, as well as manual features modeling sentiment, rhetorical structure, and POS tags, which were eventually fed into a bi-LSTM to produce IO labels, followed by some post-processing to merge neighboring spans.
For subtask TC, they extracted the pre-softmax layer of BERT and further added extra features (rhetorical, named entities, question), while taking special care of repetitions as part of a complex ensemble architecture, followed by label post-processing.

\paragraph{Team DiSaster~\cite{SemEval20-11-Kaas}\ranktc{11}} used a combination of BERT and hand-crafted features, including frequency of the fragment in the article and in the sentence it appears in and the inverse uniqueness of words in a span. The goal of the features is to compensate the inability of BERT to deal with distant context, specifically to target the technique \emph{Repetition}. 

\paragraph{Team DoNotDistribute~\cite{SemEval20-11-Kranzlein}\rankboth{22}{24}} opted for a combination of BERT-based models and engineered features (e.g., PoS, NEs, frequency within propaganda snippets in the training set). A reported performance increase of 5\% was obtained by producing 3k new silver training instances. A library was used to create near-paraphrases of the propaganda snippets by randomly substituting certain PoS words.

\paragraph{Team DUTH~\cite{SemEval20-11-Bairaktaris}\ranktc{10}} pre-processed the input including URLs normalization, number and punctuation removal, as well as lowercasing. They further mapped certain words into classes using named entity recognition with focus on person names and gazetteers containing names and variations of names of countries (255 entries), religions (35 entries), political ideologies (23 entries), slogans (41 entries). The recognized categories were replaced by the category name in the input, before passing the input to BERT.

\paragraph{Team Hitachi~\cite{SemEval20-11-Morio}\rankboth{1}{3}} used BIO encoding for subtask SI, which is typical for related segmentation and labeling tasks such as named entity recognition. They have a complex heterogeneous multi-layer neural network, trained end-to-end. The network uses a pre-trained language model, which generates a representation for each input token. To this are added part-of-speech (PoS) and named entity (NE) embeddings. As a result, there are three representations for each token, which are concatenated and used as an input to bi-LSTMs. At this moment, the network branches as it is trained with three objectives: (\emph{i})~the main BIO tag prediction objective, and two auxiliary objectives, namely (\emph{ii})~token-level technique classification, and (\emph{iii})~sentence-level classification. There is one Bi-LSTM for objectives (\emph{i})~and (\emph{ii}), and there is another Bi-LSTM for objective (\emph{iii}). For the former, there is an additional CRF layer, which helps improve the consistency of the output. 
\emph{For subtask TC}, there are two distinct FFNs, feeding input representation, which are obtained in the same manner as for subtask SI. One of the two FFNs is for sentence representation, and the other one is for the representation of tokens in the propaganda span. The propaganda span representation is obtained by concatenating representation of the begin-of-sentence token, span start token, span end token, and aggregated representation by attention and maxpooling.
\emph{For both subtasks}, these architectures are trained independently with different BERT, GPT-2, XLNet, XLM, RoBERTa, or XLM-RoBERTa; and the resulting models are combined in ensembles.

\paragraph{Team Inno~\cite{SemEval20-11-Grigorev}\ranktc{7}} used RoBERTa with cost-sensitive learning for subtask TC. They experimented with undersampling, i.e.~removing examples from the bigger classes, as well as with modeling the context. They also tried various pre-trained Transformers, but obtained worse results.

\paragraph{Team JUST~\cite{SemEval20-11-Altiti}\ranktc{15}} based its approach to task on the BERT uncased pre-trained language model, which uses 12 transformer layers that are trained for 15 epochs.

\paragraph{Team LTIatCMU~\cite{SemEval20-11-Khosla}\ranksi{4}} used a multi-granular BERT BiLSTM for subtask SI. It employs additional syntactic, semantic and pragmatic affect features at the word, sentence and document level. It has been jointly trained on token and sentence propaganda classification, with class balancing. In addition, BERT was fine-tuned to persuasive language on about 10,000 articles from propaganda websites, which turned out important in their experiments.

\paragraph{Team newsSweeper~\cite{SemEval20-11-Singh}\rankboth{13}{5}} used BERT with BIOE encoding for subtask SI.
For the TC subtask, their official run used RoBERTa to obtain representations for the snippet and for the sentence, which they concatenated.
The team further experimented (\emph{i})~with other Transformers (BERT, RoBERTa, SpanBERT, and GPT-2), (\emph{ii})~with other sequence labeling schemes (P/NP, BIO, BIOES), (\emph{iii})~with concatenating different hidden layers of BERT to obtain a token representation, and (\emph{iv})~with POS tags, as well as (\emph{v})~with different neural architectures.

\paragraph{Team NLFIIT~\cite{SemEval20-11-Martinkovic}\rankboth{17}{16}} used various combinations of neural architecture and embeddings and found out that ELMO combined with BiLSTM (and self attention for subtask TC) leads to best performances. 

\paragraph{Team NoPropaganda~\cite{SemEval20-11-Dimov}\rankboth{7}{6}} used the LasetTagger model with BERT-base encoder for the SI subtask. R-BERT was employed for the TC subtask.

\paragraph{Team NTUAAILS~\cite{SemEval20-11-Arsenos}\rankboth{27}{25}} used a residual biLSTM fed with pre-trained ELMo embeddigns for subtask SI. A biLSTM was used for subtask TC as well, this time fed with GloVe word embeddings

\paragraph{Team PsuedoProp~\cite{SemEval20-11-Chauhan}\ranksi{14}} concentrated on SI. They pre-classified sentences as being propaganda or not using an ensemble of XLNet and RoBERTa, before fine-tuning a BERT-based CRF sequence tagger to identify the exact spans. 

\paragraph{Team SkoltechNLP~\cite{SemEval20-11-Dementieva}\rankboth{25}{26}} fine-tuned BERT for SI, expanding the original training set through data augmentation techniques based on distributional semantics.

\paragraph{Team SocCogCom~\cite{SemEval20-11-Krishnamurthy}\ranktc{13}} approached subtask TC using BERT/ALBERT together with (\emph{i})~semantic-level emotional salience features from CrystalFeel~\cite{gupta-yang-2018-crystalfeel}, and (\emph{ii})~word-level psycholinguistic features from the LIWC lexicon~\cite{LIWC}. They further modeled the context, i.e.~three words before and after the target propaganda snippet.

\paragraph{Team Solomon~\cite{SemEval20-11-Raj}\ranktc{4}} addressed TC with a system that combines a transfer learning model based on fine-tuned RoBERTa (integrating fragment and context information), an ensemble of binary classifiers for the minority classes and a novel system to specifically handle \emph{Repetition}: dynamic least common sub-sequence is used to score the similarity between the fragment and the context and then the fragment is considered to be a repetition if the score is greater than a threshold heuristically set with respect to the length of the fragment. 

\paragraph{Team syrapropa~\cite{SemEval20-11-Li}\rankboth{25}{20}} fine-tuned SpanBERT, a variant of BERT for span detection, on the context of spans in terms of the surrounding non-propaganda text for subtask SI. 
For subtask TC, they use a hybrid model that consists of several submodels, each specializing in some of the relations. These models include (\emph{i})~BERT, (\emph{ii})~BERT with cost adjustment to address class imbalance, and (\emph{iii})~feature-rich logistic regression. The latter uses features such as length, TF.IDF-weighted words, repetitions, superlatives, and lists of fixed phrases targeting specific propaganda techniques. The output of the hybrid model is further post-processed using some syntactic rules based on part of speech.

\paragraph{Team Transformers~\cite{SemEval20-11-Verma}\rankboth{9}{29}}%
explored a manifold of models to address the SI subtask. They consider residual biLSTMs fed with ELMo representations as well as different variations of BERT and RoBERTa with CNNs 

\paragraph{Team TTUI~\cite{SemEval20-11-Kim}\rankboth{20}{14}} proposed an ensemble of fine-tuned BERT and RoBERTa models. They observed that feeding as input to the neural network a chunk possibly overlapping multiple sentences leads to the best performance. Moreover, for subtask SI, they applied a post-processing to remove gaps in the predictions between adjacent words. For subtask TC, they show  that the context does not increase the performance in their experiments. 

\paragraph{Team UAIC1860~\cite{SemEval20-11-Ermurachi}\rankboth{28}{26}} used traditional text representation techniques: character $n$-grams, word2vec embeddings, and TF.IDF-weighted word-based features. For both subtasks these features were used in a Random Forest classifier. Additional experiments with Na\"{i}ve Bayes, Logistic Regression and SVMs yielded worse results.

\paragraph{Team UMSIForeseer~\newcite{SemEval20-11-Jiang}\ranktc{17}} treated TC. They fine-tuned BERT on the labeled training spans, using a mix of oversampling and undersampling that is leveraged using a bagging ensemble learner. 

\paragraph{Team UNTLing~\cite{SemEval20-11-Petee}\ranktc{27}} used a logistic regression classifier for task TC, with a number of features, including bag-of-words, embeddings, NE and VAD lexicon features. Their analysis of their system highlights that NE are useful for \emph{Loaded Language} and \emph{Flag Waving}: they explain the latter since \emph{Flag Waving} usually includes references to idealized entities. The VAD features are useful for emotion-related techniques such as \emph{Appeal to fear/prejudice} and \emph{Doubt}. 
They performed some experiments on he development set for subtask SI after the deadline. They used CRF with a number of features including the PoS, the syntactic dependency of the token and of the previous/next word, BoW of preceding/following tokens and the GloVe embedding of the token.  

\paragraph{Team UPB~\cite{SemEval20-11-Paraschiv}\rankboth{5}{19}} used models based on BERT--base. Rather than just using the pre-trained models, they used masked language modeling to domain adapt it with 9M-articles with fake, suspicious, and hyperpartisan news articles. 
The same domain-adapted model was used for both subtasks. Whereas a CRF was used for the SI subtask, a softmax was used for TC.

\paragraph{Team UTMN~\cite{SemEval20-11-Mikhalkova}\ranksi{23}} addressed the SI subtask by representing the texts with a concatenation of tokens and context embeddings, together with sentiment intensity from VADER. They avoided deep learning architectures in order to produce a 
computationally-affordable model and opted for a logistic regressor instead.

\paragraph{Team WMD~\cite{SemEval20-11-Daval-Frerot}\rankboth{33}{21}} used an ensemble of  BERT-based neural models, LSTMs, SVMs, gradient boosting and random forest together with character and word-level embeddings. In addition, they use a number of techniques for data augmentation: back-translation, synonym replacement and TF-IDF replacement (replace unimportant words, according to their TF-IDF score, with other unimportant words). 

\paragraph{Team YNU-HPCC~\cite{SemEval20-11-Dao}\rankboth{16}{22}} participated in both subtasks using Glove and BERT embeddings in combination with LSTM, BiLSTM and XGBoost. 

\paragraph{Team YNUtaoxin~\cite{SemEval20-11-Tao}\ranksi{11}} tested BERT, RoBERTa and XLNet on the SI subtask focusing on determining the optimal input sentence length for the networks.

\newpage
\section{Errata Corrige}
\label{app:errata}

After the end of the shared task, a bug in the evaluation functions was found. 
The bug affected both the SI and the TC tasks. Still, it had a low impact  and the ranking computed with the fixed software does not change significantly ---in particular for the top-ranked submissions. 
Tables~\ref{tab:span-test-errata} and~\ref{tab:technique-test-errata} show the correct scores on the test sets for tasks SI and TC. 
Any reference to the task results should refer to  these numbers. 

\newcolumntype{G}{>{\columncolor{Gray}}r}
\begin{table}[H]
\centering

\begin{tabular}{l|Gggg}
\hline 
& \multicolumn{4}{|c}{\bf Test}	\\
\rowcolor{white}
\bf Team  & \bf Rnk & \bf F$_1$		& \bf P	& \bf R	\\ \hline
\rowcolor{white}
Hitachi		& 1& \bf  51.74		& 55.76	& 48.27	\\
ApplicaAI	& 2	& 49.88		& 59.33	& 43.02	\\
\rowcolor{white}
aschern		& 3	& 49.59		& 52.57	& 46.93	\\
LTIatCMU	& 4	& 48.16		& 50.35	& 46.15	\\
\rowcolor{white}
UPB		& 5	& 46.63	& 57.50	& 39.22	\\
Fragarach	& 6	& 46.43		& 53.44	& 41.05	\\
\rowcolor{white}
NoPropaganda	& 7	& 45.17		& 55.05	& 38.29	\\
YNUtaoxin	& 8	& 43.80		& 54.60	& 36.57	\\
\rowcolor{white}
Transformers	& 9	& 43.77		& 49.05	& 39.52 \\
SWEAT		& 10	& 43.69		& 52.13	& 37.61	\\
\rowcolor{white}
DREAM		& 11	& 43.60		& 53.67	& 36.71	\\
CyberWallE	& 12	& 43.59		& 40.99	& 46.54	\\
\rowcolor{white}
newsSweeper	& 13	& 42.20		& 45.30	& 39.49	\\
PsuedoProp	& 14	& 41.81		& 41.24	& 42.41	\\
\rowcolor{white}
Solomon		& 15	& 41.26		& 53.69	& 33.51	\\
NLFIIT		& 16	& 41.10		& 50.17	& 34.81	\\
\rowcolor{white}
TTUI		& 17	& 40.76	    & \bf 66.37	& 29.41	\\
PALI		& 18	& 40.73		& 52.10	& 33.44	\\
\rowcolor{white}
YNUHPCC		& 19	& 40.46		& 36.35	& 45.63	\\
UESTCICSA	& 20	& 40.41		& 55.58	& 31.74	\\
\rowcolor{white}
BPGC		& 21	& 38.89		& 48.50	& 32.45	\\
DoNotDistribute	& 22	& 37.92		& 41.47	& 34.92	\\
\rowcolor{white}
UTMNandOCAS	& 23	& 37.71		& 37.12	& 38.31	\\
Entropy		& 24	& 37.31		& 40.82	& 34.35	\\
\rowcolor{white}
syrapropa	& 25	& 36.92		& 49.22	& 29.53	\\
SkoltechNLP	& 26	& 34.36		& 45.77	& 27.51	\\
\rowcolor{white}
NTUAAILS	& 27	& 34.36		& 45.62	& 27.55	\\
UAIC1860	& 28	& 32.67		& 23.86	& 51.78	\\
\rowcolor{white}
CCNI		& 29	& 29.68		& 37.73	& 24.46	\\
NCCU-SMRG	& 30	& 27.69		& 16.70	& \bf 80.94	\\
\rowcolor{white}
3218IR		& 31	& 23.28		& 21.95	& 24.79	\\
WMD		& 32	& 20.51		& 45.44	& 13.24	\\
\rowcolor{white}
LS		& 33	& 18.14		& 33.20	& 12.47	\\
HunAlize	& 34	& \,\,3.15	& 22.39	& \,\,1.69 \\
\rowcolor{white}
YOLO		& 35	& \,\,0.74	& 18.32	& \,\,0.38 \\
Baseline	& 36	& \,\,0.32	& 13.04	& \,\,0.16 \\
\hline
\end{tabular}

\caption{\label{tab:span-test-errata} \textbf{Subtask 1: Span Identification (SI) performance on test after bug fixing.} The highest scores for the different measures appear highlighted.}
\end{table}

\begin{landscape}
\begin{table}
\renewcommand{\arraystretch}{0.94}%
\begin{tabular}{rl| G |GGGGGGG GGGGGGG}
\rowcolor{white}
Rnk & Team	& Overall	& 1	& 2	& 3	& 4	& 5	& 6	& 7	& 8	& 9	& 10	& 11	& 12	& 13	& 14	\\
\hline 
\rowcolor{white}
1 & ApplicaAI & \bf 63.74 & 48.15 & \bf 47.06 & 8.33 & 50.98 & 22.73 & 56.23 & 36.64 & \bf 70.47 & 78.27 & 75.80 & \bf 58.79 & \bf 58.97 & 39.22 & 28.13  \\
2 & aschern & 63.30 & 35.29 & 41.78 & 14.93 & \bf 53.57 & 25.97 & 59.44 & 37.55 & 66.35 & \bf 78.32 & \bf 76.68 & 56.94 & 54.24 & \bf 42.55 & 22.64  \\
\rowcolor{white}
3 & Hitachi & 63.13 & 40.00 & 38.94 & 4.88 & 47.92 & \bf 36.62 & \bf 63.43 & 37.38 & 69.04 & 76.69 & 76.22 & 42.42 & 52.63 & 29.41 & 26.92  \\
4 & NoPropaganda & 59.83 & 23.26 & 38.02 & 0.00 & 8.70 & 12.12 & 56.27 & \bf 39.43 & 60.30 & 77.99 & 76.61 & 48.14 & 51.52 & 27.91 & 0.00  \\
\rowcolor{white}
5 & Solomon & 59.39 & 26.67 & 39.39 & 4.88 & 42.00 & 19.61 & 61.82 & 32.00 & 61.84 & 75.30 & 70.75 & 42.53 & 50.75 & 38.10 & 0.00 \\
6 & CyberWallE & 58.99 & 14.63 & 39.84 & 12.25 & 26.32 & 15.39 & 59.24 & 33.20 & 56.22 & 75.81 & 71.60 & 52.98 & 45.46 & 23.81 & 0.00  \\
\rowcolor{white}
7 & newsSweeper & 58.44 & 41.67 & 42.80 & 13.04 & 25.00 & 19.72 & 56.55 & 37.10 & 60.53 & 75.32 & 74.23 & 20.69 & 50.75 & 21.62 & 8.00  \\
8 & Inno & 57.99 & 43.14 & 45.09 & 10.71 & 40.40 & 24.32 & 58.65 & 35.39 & 59.41 & 73.31 & 74.30 & 24.89 & 43.75 & 29.63 & 19.36  \\
\rowcolor{white}
9 & djichen & 57.71 & 37.50 & 34.58 & 20.83 & 43.81 & 22.79 & 60.00 & 32.18 & 56.19 & 73.99 & 70.43 & 36.36 & 30.77 & 27.91 & 18.87  \\
10 & PALI & 57.43 & 27.59 & 36.59 & \bf 28.57 & 39.62 & 30.59 & 58.97 & 28.57 & 61.62 & 74.29 & 69.43 & 25.44 & 39.22 & 40.82 & 20.90  \\
\rowcolor{white}
11 & DUTH & 57.21 & 44.44 & 33.33 & 9.30 & 33.33 & 26.23 & 59.16 & 28.24 & 58.95 & 73.71 & 71.41 & 20.10 & 34.78 & 27.03 & 17.78  \\
12 & DiSaster & 56.65 & \bf 51.16 & 35.25 & 20.41 & 26.67 & 14.63 & 59.12 & 30.64 & 58.25 & 74.49 & 68.10 & 20.44 & 42.55 & 19.05 & 4.35  \\
\rowcolor{white}
13 & SocCogCom & 55.81 & 28.57 & 31.62 & 9.76 & 26.51 & 6.35 & 60.40 & 34.86 & 54.26 & 72.18 & 67.34 & 18.88 & 40.91 & 23.53 & 10.00  \\
14 & TTUI & 55.64 & 28.24 & 41.48 & 24.56 & 35.29 & 23.08 & 57.40 & 32.20 & 61.68 & 73.22 & 68.49 & 21.18 & 37.50 & 25.00 & 20.29  \\
\rowcolor{white}
15 & JUST & 55.36 & 48.89 & 37.10 & 24.49 & 31.11 & 27.27 & 58.26 & 29.57 & 62.56 & 71.96 & 64.73 & 21.94 & 33.33 & 31.82 & \bf 28.57  \\
16 & NLFIIT & 55.25 & 29.63 & 24.89 & 9.30 & 32.10 & 0.00 & 55.66 & 30.30 & 63.32 & 72.55 & 69.30 & 21.55 & 41.67 & 13.64 & 0.00  \\
\rowcolor{white}
17 & UMSIForeseer & 55.14 & 29.41 & 31.97 & 9.76 & 0.00 & 0.00 & 57.21 & 28.57 & 56.14 & 73.02 & 70.79 & 21.49 & 39.22 & 14.29 & 0.00  \\
18 & BPGC & 54.81 & 30.77 & 33.65 & 12.50 & 40.00 & 24.18 & 53.78 & 33.47 & 58.93 & 71.58 & 67.51 & 23.74 & 40.00 & 20.69 & 20.90  \\
\rowcolor{white}
19 & UPB & 54.30 & 20.00 & 30.00 & 0.00 & 19.72 & 16.95 & 52.55 & 30.62 & 55.87 & 70.09 & 68.86 & 20.00 & 34.62 & 22.86 & 4.88 \\
20 & syrapropa & 54.25 & 31.58 & 29.77 & 19.61 & 30.61 & 21.51 & 56.14 & 28.10 & 57.02 & 71.47 & 68.44 & 30.77 & 29.03 & 28.57 & 9.09 \\
\rowcolor{white}
21 & WMD & 52.35 & 7.14 & 29.20 & 5.00 & 0.00 & 5.71 & 54.42 & 26.36 & 52.08 & 69.76 & 64.67 & 14.82 & 6.90 & 7.41 & 0.00  \\
22 & YNUHPCC & 50.50 & 29.79 & 26.40 & 13.79 & 38.38 & 3.45 & 51.04 & 21.54 & 55.56 & 68.08 & 62.33 & 17.72 & 27.59 & 17.78 & 15.00  \\
\rowcolor{white}
23 & UESTCICSA & 49.94 & 12.77 & 22.75 & 0.00 & 11.27 & 3.74 & 44.99 & 25.44 & 53.14 & 68.23 & 66.88 & 27.96 & 41.38 & 28.57 & 3.70  \\
24 & DoNotDistribute & 49.72 & 22.64 & 29.75 & 4.55 & 24.49 & 14.89 & 46.25 & 27.23 & 53.76 & 68.44 & 60.65 & 19.44 & 28.07 & 12.25 & 9.68  \\
\rowcolor{white}
25 & NTUAAILS & 46.37 & 36.36 & 28.31 & 9.76 & 0.00 & 0.00 & 48.75 & 29.66 & 46.47 & 65.79 & 54.55 & 18.43 & 13.79 & 11.43 & 4.08  \\
26 & UAIC1860 & 41.17 & 17.39 & 22.12 & 9.76 & 2.90 & 7.60 & 36.41 & 21.01 & 38.78 & 62.33 & 42.97 & 11.16 & 11.43 & 5.56 & 4.26  \\
\rowcolor{white}
27 & UNTLing & 39.11 & 4.35 & 5.29 & 0.00 & 0.00 & 2.86 & 32.65 & 11.82 & 40.48 & 62.57 & 36.74 & 7.78 & 17.65 & 0.00 & 0.00  \\
28 & HunAlize & 37.10 & 6.45 & 11.83 & 0.00 & 8.00 & 7.84 & 31.76 & 23.81 & 29.95 & 58.59 & 15.82 & 2.09 & 4.55 & 0.00 & 0.00  \\
\rowcolor{white}
29 & Transformers & 26.54 & 0.00 & 0.00 & 0.00 & 0.00 & 0.00 & 0.98 & 0.00 & 0.00 & 47.55 & 24.06 & 2.86 & 0.00 & 0.00 & 0.00  \\
30 & Baseline & 25.20 & 0.00 & 3.68 & 0.00 & 0.00 & 11.56 & 29.14 & 14.42 & 6.20 & 46.48 & 0.00 & 19.26 & 0.00 & 0.00 & 0.00 \\
\rowcolor{white}
31 & IJSE8 & 20.62 & 0.00 & 7.07 & 0.00 & 0.00 & 2.27 & 17.07 & 10.70 & 8.57 & 39.32 & 15.39 & 4.92 & 0.00 & 0.00 & 0.00  \\
32 & Entropy & 20.50 & 3.92 & 7.12 & 0.00 & 2.27 & 7.41 & 12.81 & 6.39 & 4.95 & 38.02 & 15.14 & 5.83 & 0.00 & 0.00 & 6.78  \\
\end{tabular}
\caption{\label{tab:technique-test-errata} \textbf{Technique classification F$_1$ performance on the test set after bug-fixing}. The systems are ordered on the basis of the final 
ranking. Columns \textbf{1} to \textbf{14} show the performance on each of the 
fourteen classes (cf.\ Section~\ref{sec:propagandatechniques}). The best score 
on each technique appears highlighted.}
\end{table}
\end{landscape}

\newpage
\section{Annotation Instructions}
\label{app:annotation-instructions}

We report below a series of snapshots of the annotation instructions. 

\begin{figure}[htb]
    \centering
    \includegraphics[width=0.85\columnwidth]{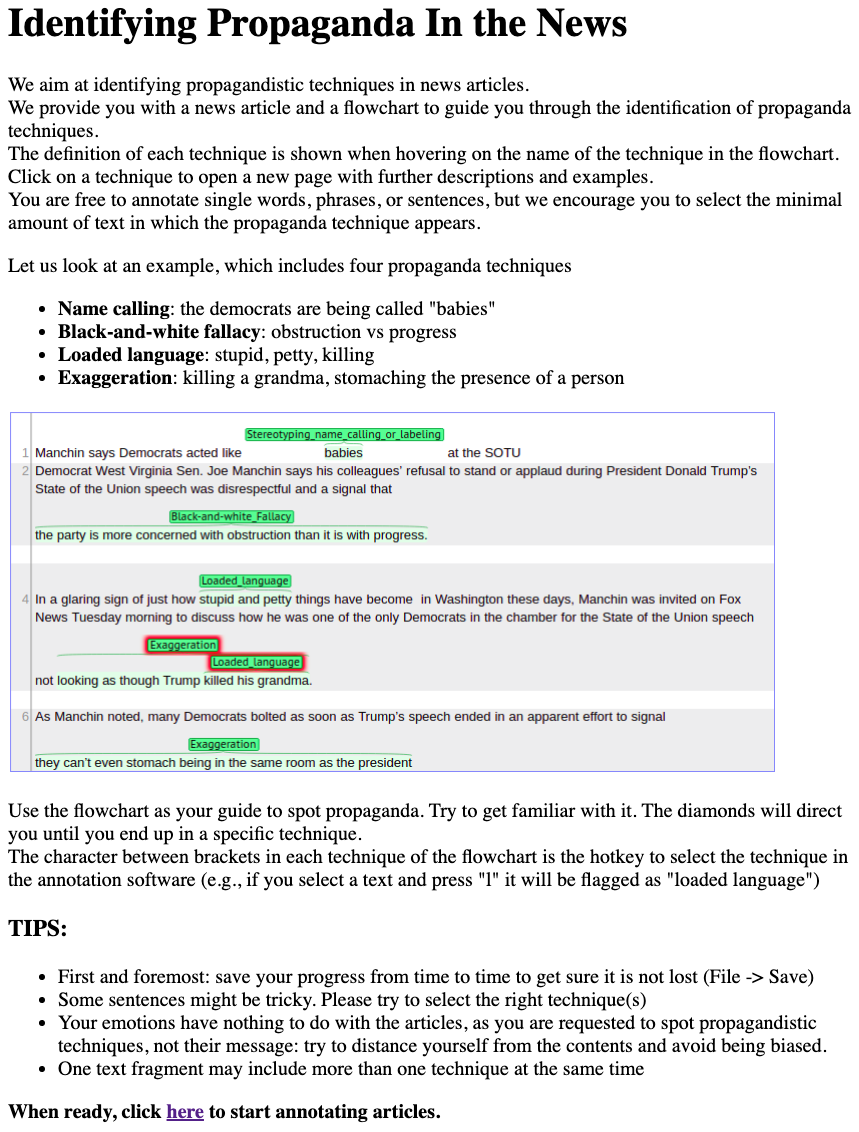}
    \caption{Instruction for the annotators.}
    \label{fig:annotation:instructions}
\end{figure}

\begin{figure}[htb]
    \centering
    \includegraphics[width=0.8\columnwidth]{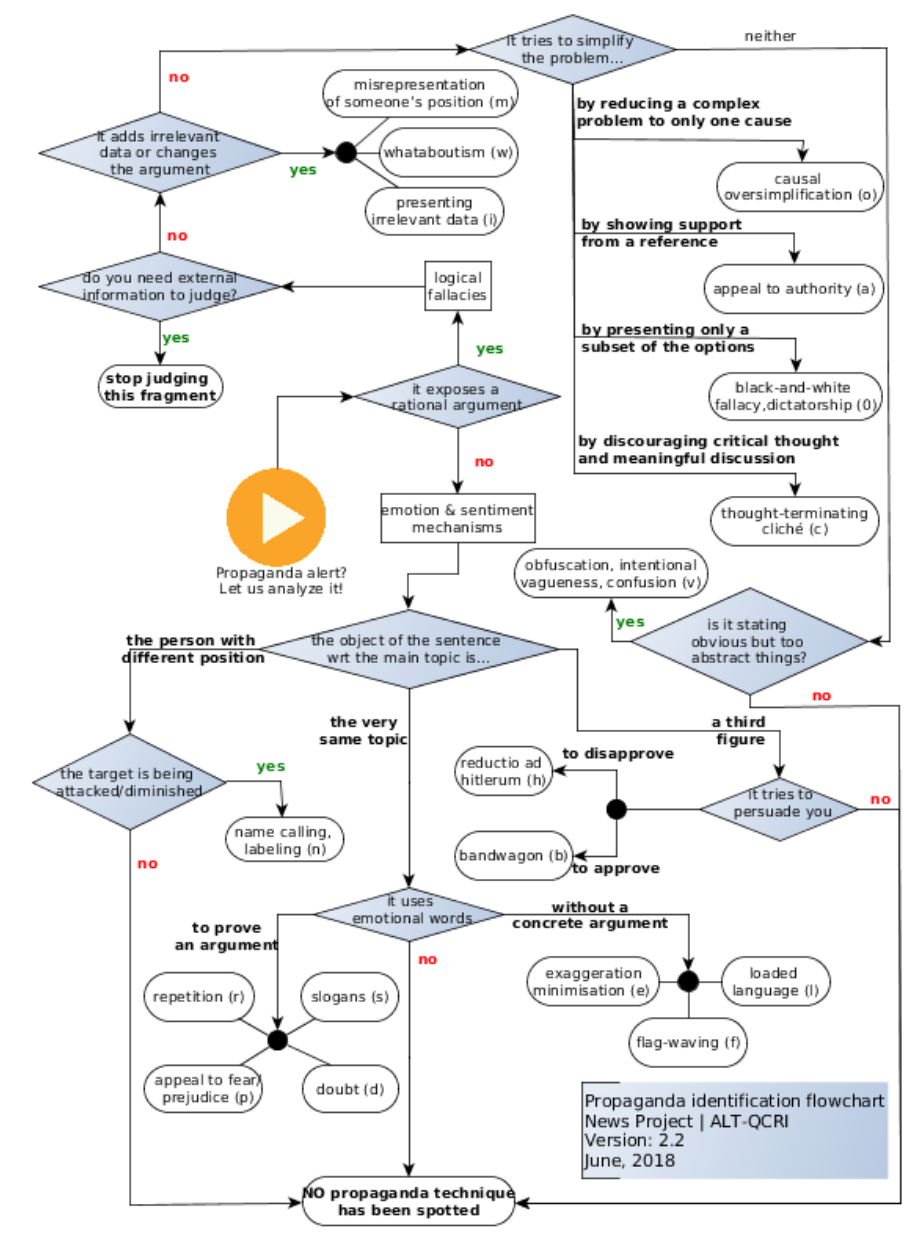}
    \caption{Annotation instructions: hierarchical diagram to guide the choice of technique.}
    \label{fig:annotation:diagram}
\end{figure}

\begin{figure}[htb]
    \centering
    \includegraphics[width=0.9\columnwidth]{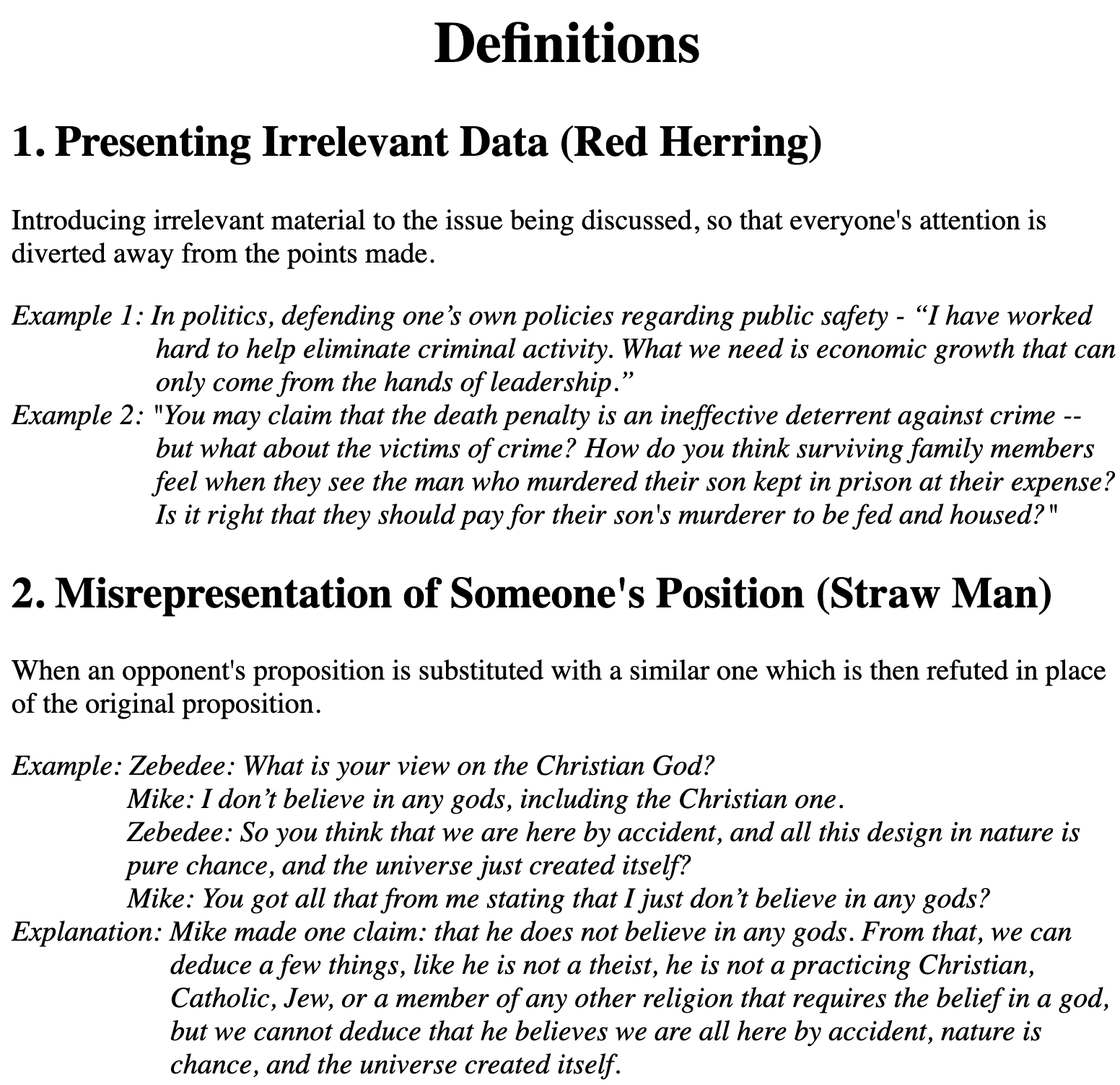}
    \label{fig:annotation:instr1}
\end{figure}

\begin{figure}[htb]
    \centering
    \includegraphics[width=0.9\columnwidth]{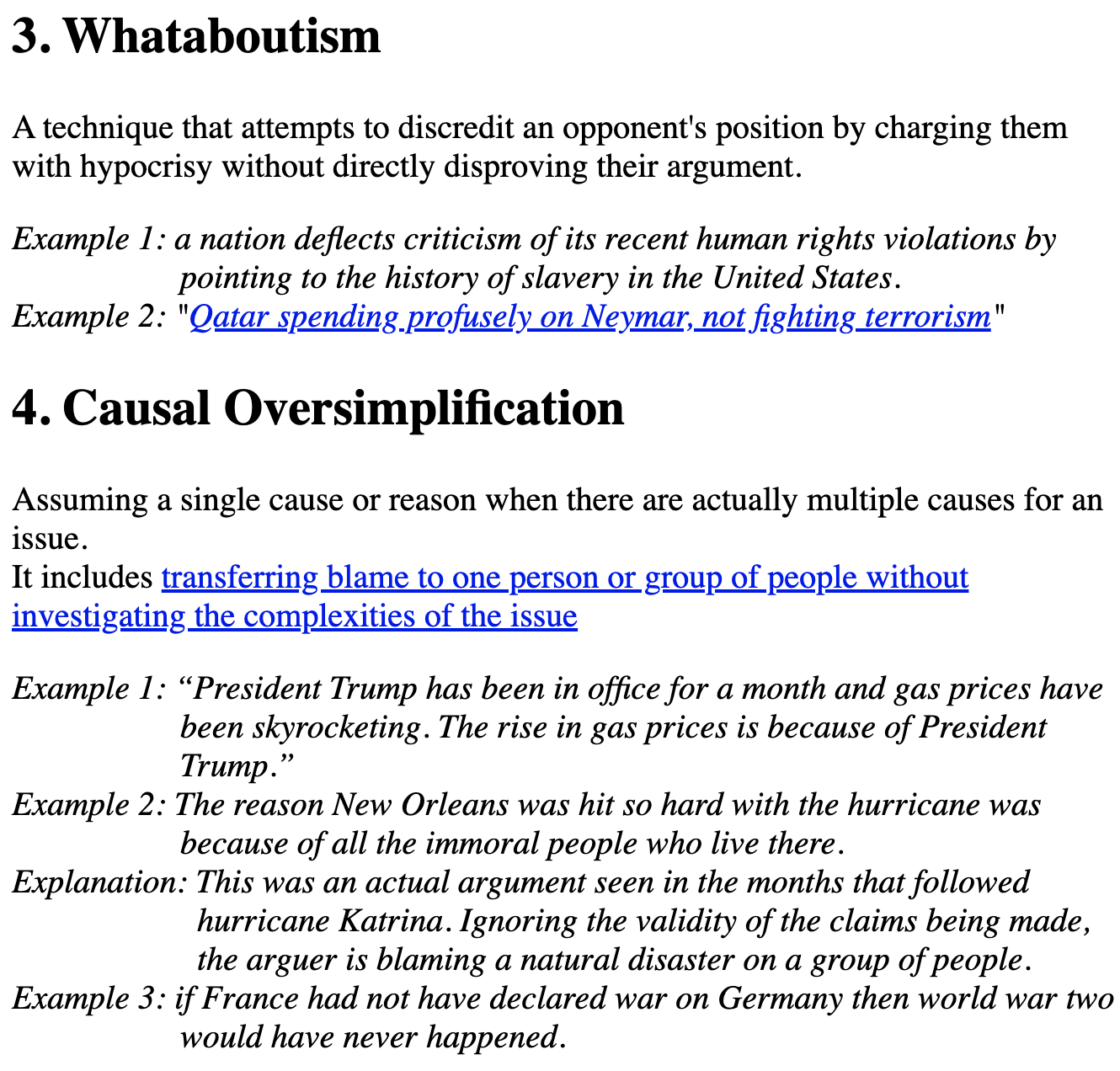}
    \label{fig:annotation:instr2}
\end{figure}

\begin{figure}[htb]
    \centering
    \includegraphics[width=0.9\columnwidth]{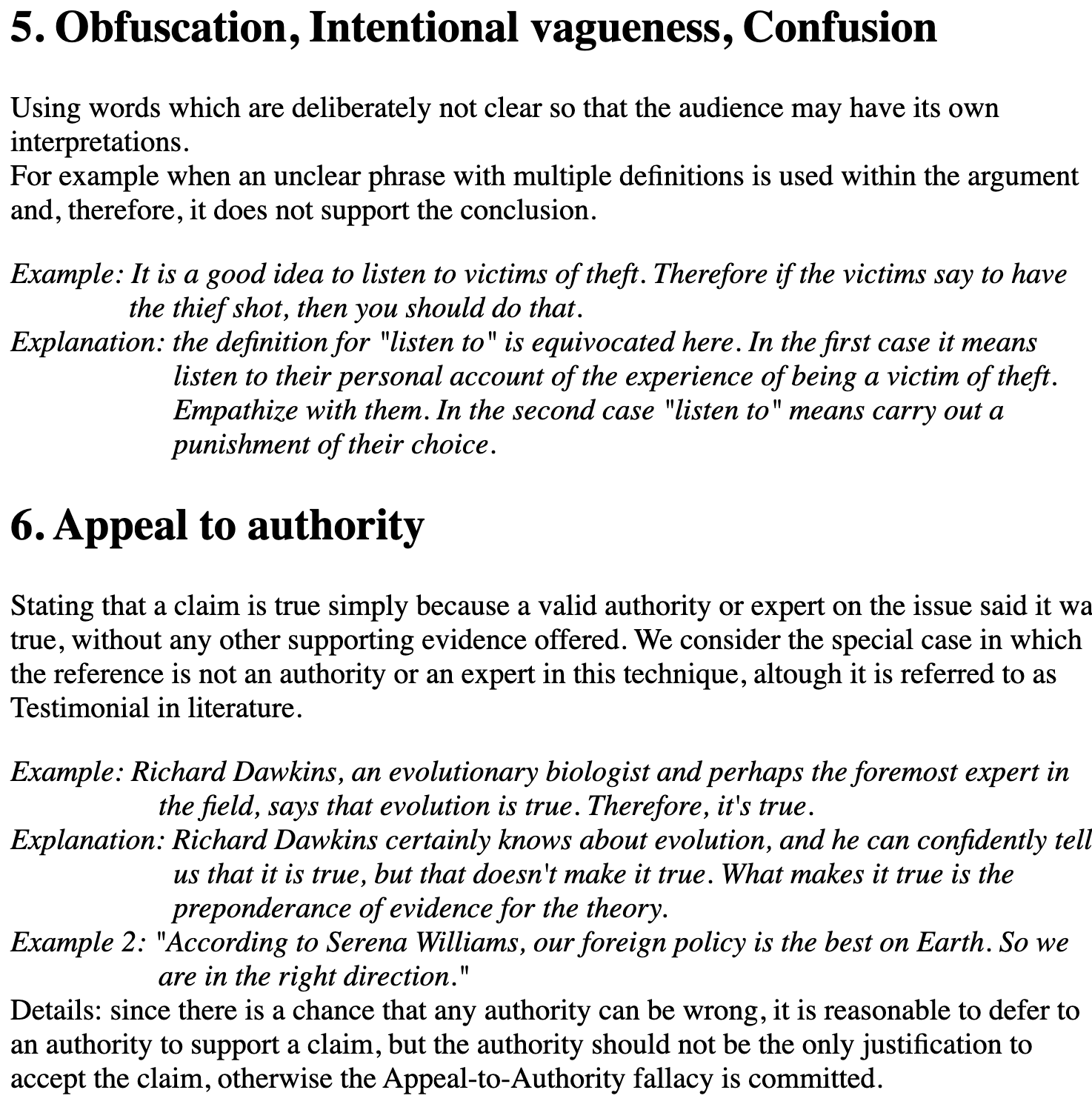}
    \label{fig:annotation:instr3}
\end{figure}

\begin{figure}[htb]
    \centering
    \includegraphics[width=0.9\columnwidth]{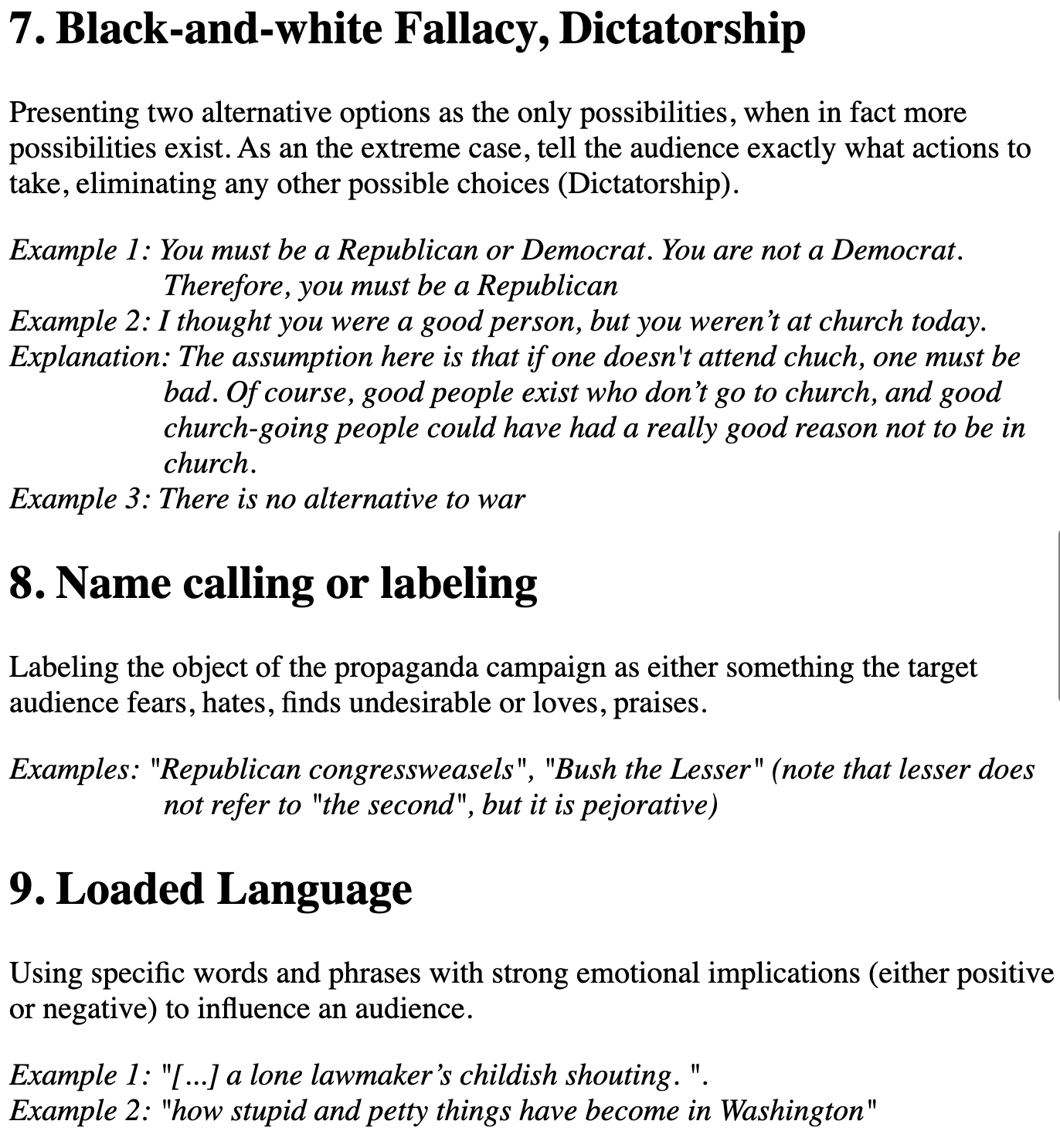}
    \label{fig:annotation:instr4}
\end{figure}

\begin{figure}[htb]
    \centering
    \includegraphics[width=0.9\columnwidth]{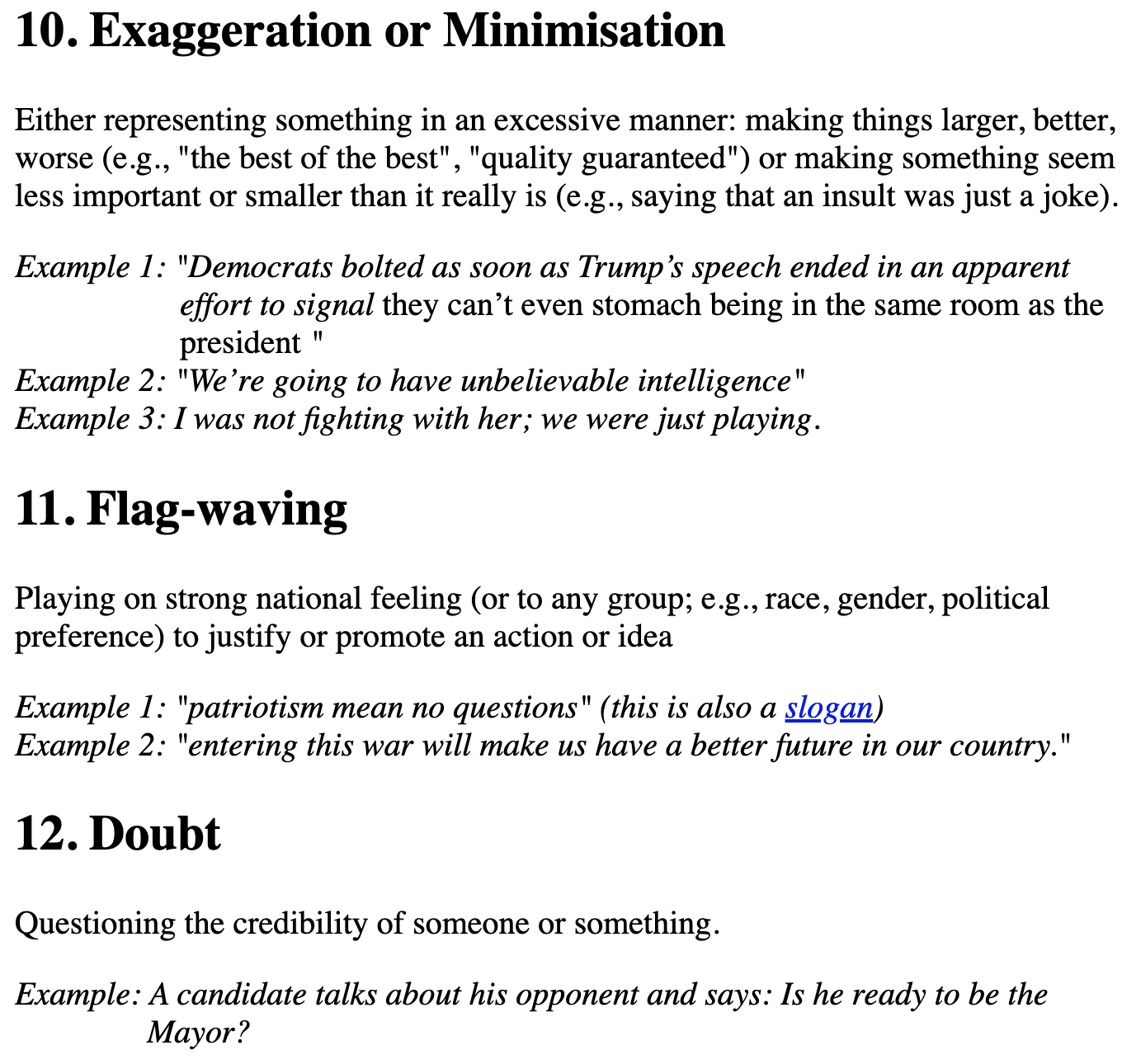}
    \label{fig:annotation:instr5}
\end{figure}

\begin{figure}[htb]
    \centering
    \includegraphics[width=0.9\columnwidth]{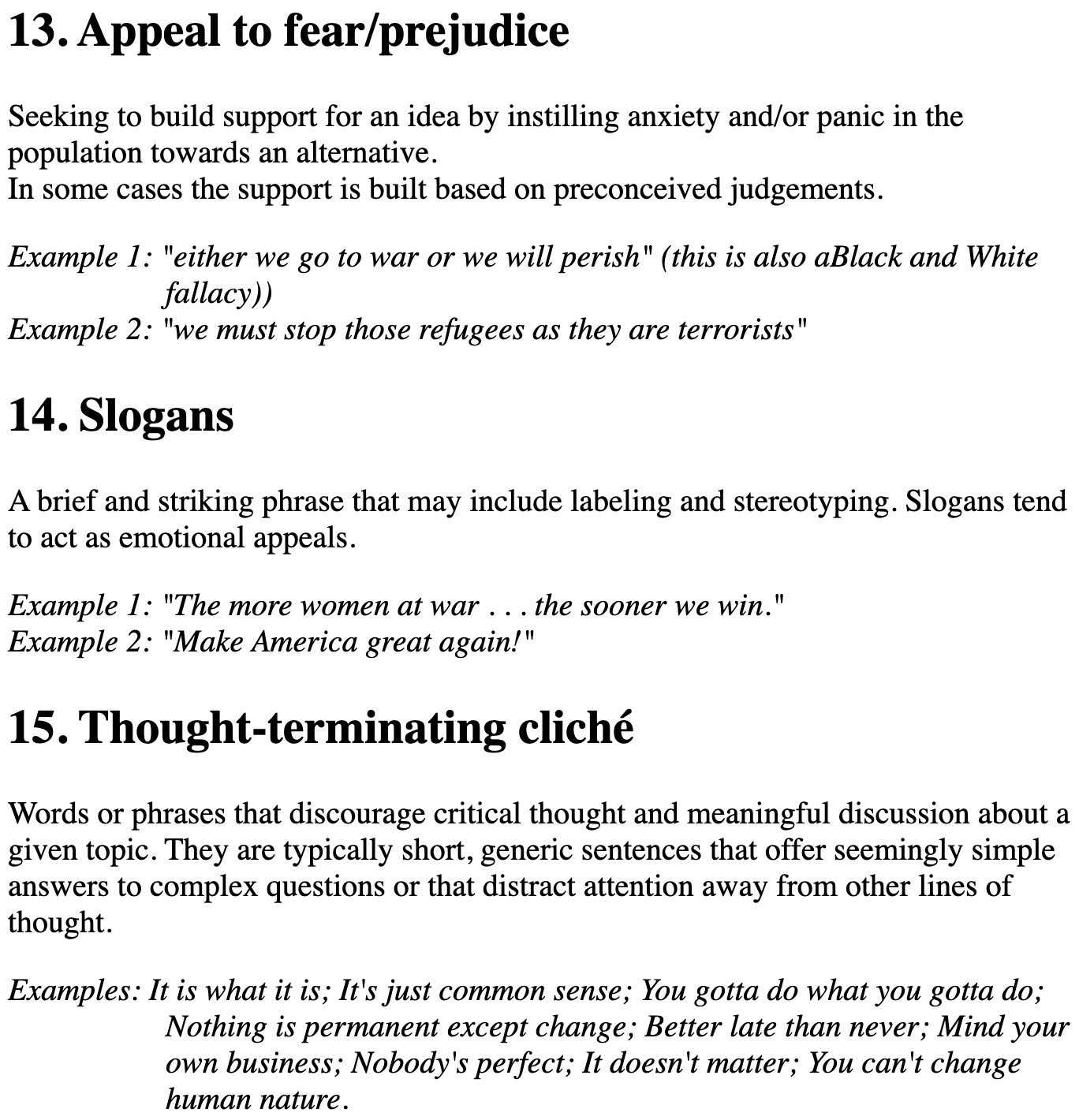}
    \label{fig:annotation:instr6}
\end{figure}

\begin{figure}[htb]
    \centering
    \includegraphics[width=0.8\columnwidth]{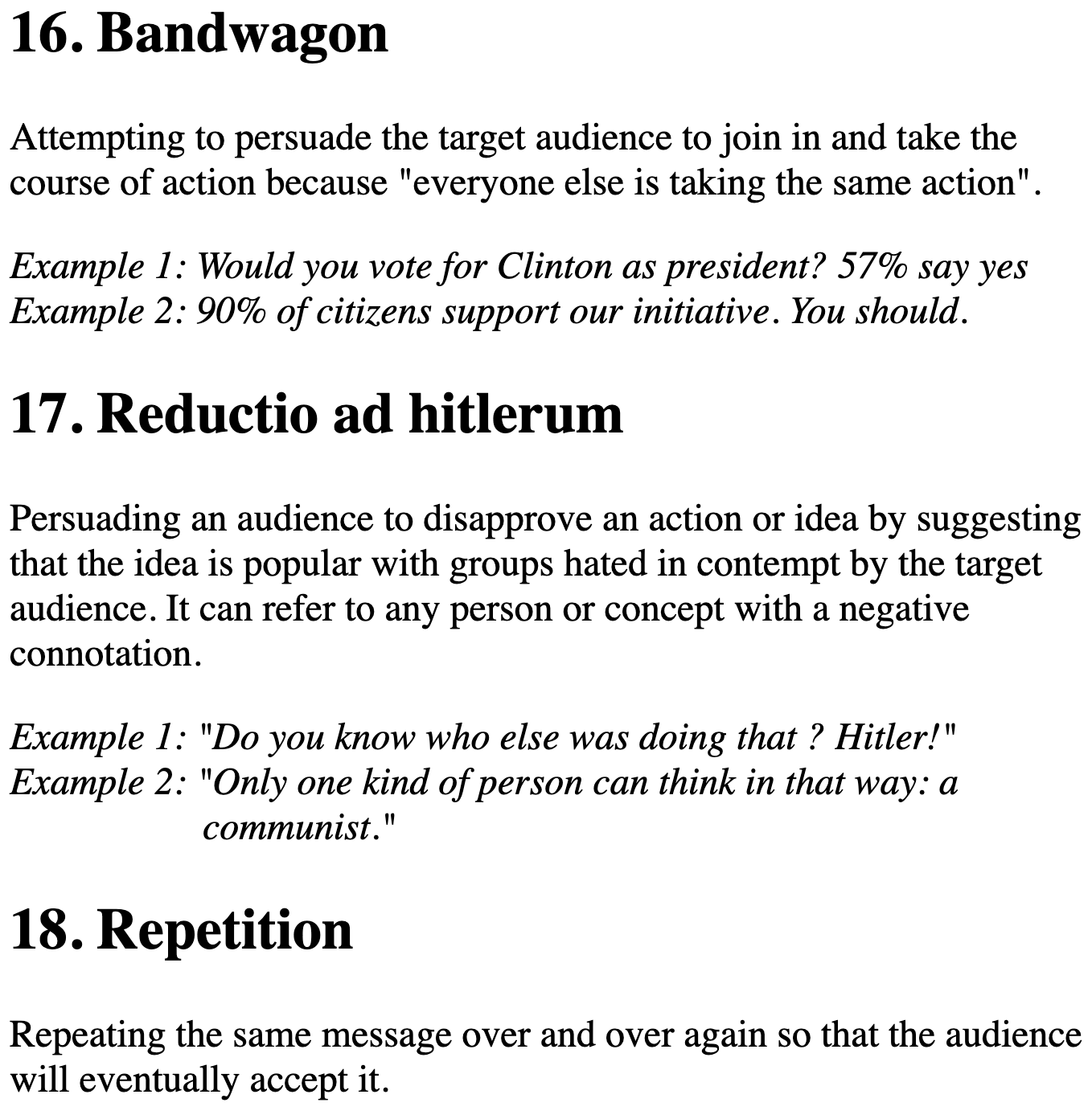}
    \label{fig:annotation:instr7}
\end{figure}

\end{document}